\documentclass[acmlarge]{acmart}

\makeatletter
\newcommand{\confshort}{\acmConference@shortname}
\newcommand{\conffull}{\acmConference@name}
\newcommand{\confdate}{\acmConference@date}
\newcommand{\confloc}{\acmConference@venue}
\AtBeginDocument{
  \fancypagestyle{firstpagestyle}{
    \fancyhead{}%
    \fancyfoot[C]{}%
  }
  \fancyhf{}
  \fancyhead[LO]{\@headfootfont\shorttitle}%
  \fancyhead[RE]{\@headfootfont\@shortauthors}%
  \fancyhead[LE]{\@headfootfont\footnotesize \confshort, \confdate, \confloc}%
  \fancyhead[RO]{\@headfootfont\footnotesize \confshort, \confdate, \confloc}%
  \fancyfoot[C]{}%
}
\makeatother
\acmBooktitle{\conffull\@ (\confshort), \confdate, \confloc}

\usepackage{booktabs}     
\usepackage{multirow}
\usepackage{threeparttable}
\usepackage{enumitem}
\usepackage{bbm} 
\usepackage{float}
\usepackage{subcaption}
\usepackage{array}
\usepackage{comment}
\usepackage{listings}
\usepackage{xcolor}
\setcopyright{acmlicensed}
\copyrightyear{2026}
\acmYear{2026}
\acmDOI{10.1145/3805689.3812264}
\acmBooktitle{The 2026 ACM Conference on Fairness, Accountability, and Transparency (FAccT '26), June 25--28, 2026, Montreal, QC, Canada}
\acmConference[FAccT '26]{The 2026 ACM Conference on Fairness, Accountability, and Transparency}{June 25--28, 2026}{Montreal, Canada}
\acmISBN{979-8-4007-2596-8/2026/06}

\begin{document}

\title{Are LLMs Ready for Conflict Monitoring? Empirical Evidence from West Africa}

\author{Hoffmann Muki}
\affiliation{%
  \institution{University of Maryland}
  \city{College Park}
  \country{USA}
}
\email{hoffmuki@umd.edu}

\author{Olukunle Owolabi}
\affiliation{%
  \institution{Independent Researcher}
  \city{Seattle}
  \country{USA}
}
\email{olukunle.owolabi@alumni.tufts.edu}

\renewcommand{\shortauthors}{Hoffmann and Kunle}

\begin{abstract}
As LLMs enter conflict monitoring, understanding systematic distortions in their outputs is critical for humanitarian accountability. We evaluate four vanilla open-weight models Gemma 3 4B, Llama 3.2 3B, Mistral 7B, and OLMo 2 7B and two domain-adapted models, AfroConfliBERT and AfroConfliLLAMA, on Nigeria and Cameroon conflict-event classification against ACLED, a gold-standard dataset with multi-stage verification. We find a bifurcated divergence in normative directionality. Open-weight models exhibit statistically significant False Illegitimation bias: Gemma misclassifies to 18.29\% of legitimate battles as civilian-targeted violence while making zero False Legitimation errors. By contrast, AfroConfliBERT and AfroConfliLLAMA achieve near-directional neutrality, with Legitimization Bias differences indistinguishable from zero. Yet domain adaptation does not eliminate actor-based selection bias. Both adapted models show statistically significant actor bias comparable to vanilla LLMs; in Nigeria, state actors are legitimized 36.5\% more often than non-state actors in identical tactical contexts. Open-weight outputs are also fragile to geography-specific lexical framing: delegitimizing phrases produce flip rates up to 66.7\% in Cameroon and 34.2\% in Nigeria, while perturbations salient in one context may not matter in another. Error trace profiling shows models mask normative bias through unfaithful rationale confabulations. In contrast, AfroConfliBERT and AfroConfliLLAMA are largely robust, with near-zero flip rates across perturbation categories. Overall, current models are not ready for unsupervised deployment in conflict monitoring. We call for fairness-aware fine-tuning to reduce actor-based selection bias, mandatory adversarial robustness evaluation against lexical manipulation, and context-specific human-in-the-loop oversight calibrated to regional difficulty.
\end{abstract}



\maketitle

\section{Introduction}
\label{sec:introduction}

Growing interest in applying Large Language 
Models (LLMs) to humanitarian, security, and policy-making contexts has prompted recent exploration of their potential as analytical engines for automated conflict event extraction \cite{Nemkova2025, Sanchez2025}. As this exploration grows, so does their potential role in the decision-making pipelines of non-governmental organizations and international security bodies. Yet the application of LLMs to conflict monitoring raises critical questions about their normative neutrality: whether model outputs faithfully reflect the empirical realities of political violence or introduce systematic distortions that could undermine the credibility of human rights documentation and humanitarian response \cite{IFIT2025, Nemkova2025, Sanchez2025}.

The utility of any automated monitoring system in this domain depends on its adherence to a rigorous, operationalizable conflict taxonomy. This study utilizes the Armed Conflict Location \& Event Data (ACLED) project as our foundational ground truth. ACLED is widely regarded as the gold standard for near-real-time conflict monitoring due to its independent, impartial, and highly granular coding methodology, which provides the necessary technical distinctions required for legal and security analysis \cite{Raleigh2023}.

Critically, the classification of conflict events is not merely a linguistic task but a normative one. Prior work demonstrates that LLM outputs vary systematically with the framing and language used to describe conflict events \cite{Steinert2025}, and that alignment procedures may introduce representational asymmetries across geographic and demographic contexts \cite{Ryan2024}. However, how such patterns manifest when LLMs are applied to conflict data, whether they vary systematically across model architectures, and whether domain adaptation can mitigate them, have not been addressed heretofore. This study addresses those questions through a meticulous empirical investigation of general-purpose and domain-adapted models on conflict event data from Nigeria and Cameroon, via equalized odds, counterfactual analysis, few-shot prompting and error trace profiling.

We investigate these questions using conflict event data from Nigeria and Cameroon, evaluating four general-purpose LLMs: Gemma 3 4B, Llama 3.2 3B, Mistral 7B, and OLMo 2 7B, alongside two domain-adapted models introduced in this study: AfroConfliBERT, a domain-adapted version of ConfliBERT fine-tuned on 81,104 ACLED African conflict events, and AfroConfliLLAMA, a Llama-based decoder fine-tuned on the same corpus via LoRA. For the sake of brevity, we will henceforth refer to the vanilla LLMs as Gemma, Llama, Mistral and Olmo respectively.

Our primary contributions are five-fold:
\begin{itemize}
    \item We provide the first systematic comprehensive evaluation 
    of LLM bias on ACLED, bridging the gap between computational linguistics and empirical conflict studies, through a targeted analysis of Cameroon and Nigeria.
    
    \item We offer a rigorous quantification of normative directionality in model errors, examining whether observed divergences across 
    architectures reflect systematic actor-based bias.
    
    \item We evaluate the effect of few-shot in-context learning (ICL) and lexical framing on vanilla LLMs, examining whether few-shot examples and lexical cues shift normative directionality in model outputs, or whether such factors merely redistribute rather than eliminate underlying bias.

    \item We perform a granular error-trace analysis using Rationale-Flip Concordance ~\cite{ye2022unreliability} and Layer Integrated Gradients ~\cite{sundararajan2017axiomatic}, demonstrating that apparent gains in model calibration often mask underlying confabulation, where models justify noteworthy flips through untrustworthy explanations.
    
    \item We introduce AfroConfliBERT and AfroConfliLLAMA, domain-adapted versions of ConfliBERT and Llama fine-tuned on ACLED African conflict event data, investigating whether domain adaptation mitigates the observed biases and what limitations remain.
\end{itemize}

The remainder of this paper is organized as follows. Section~\ref{sec:background} situates our work within the socio-technical literature on LLMs in conflict analysis, followed by a detailed account of our methodology and experimental design in Sections~\ref{sec:methodology} and \ref{sec:experimental_design}. Section~\ref{sec:results} presents our empirical findings and their interpretation, while Section~\ref{sec:conclusion} concludes by discussing the implications for the responsible deployment of AI in international security policy. 

To provide further technical depth, the Appendix includes per-class model metrics (Section~\ref{appendix:per-class_model_metrics}), formal definitions for legitimization and fairness (Sections~\ref{appendix:metrics} and \ref{appendix:fairness_metrics}), and an analysis of model calibration (Section~\ref{appendix:calibration_metrics}). We also examine model sensitivity to contextual length in Section~\ref{appendix:error_correlation_with_text_length} and conclude with a qualitative audit of the primary model disagreements in Section~\ref{appendix:qualitative_audit_of_disagreements}.

\begin{figure}[t]
    \centering
    \includegraphics[width=\linewidth]{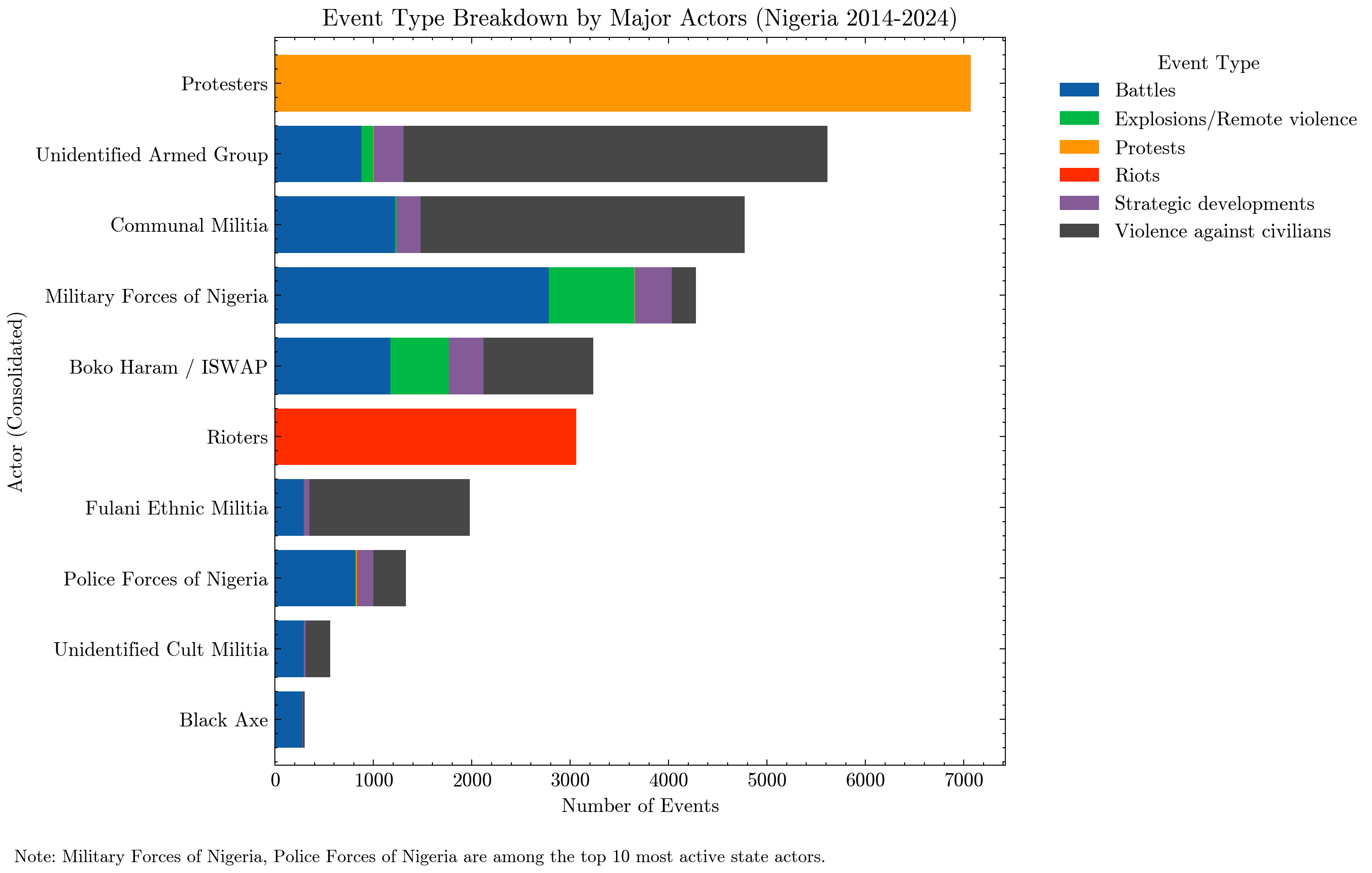}
    \caption{Top ten conflict actors in Nigeria (2014–2024), illustrating the diverse landscape of conflict with participation from state actors, community militias, and other non-state groups.}
    \Description[Top ten conflict actors in Nigeria]{Top ten conflict actors in Nigeria (2014–2024), illustrating the diverse landscape of conflict with participation from state actors, community militias, and other non-state groups.}
    \label{fig:conflict_actors_barchart}
\end{figure}


\section{Background and Related Work}
\label{sec:background}

\paragraph{Actor Definitions}
Analysis of political violence requires a precise taxonomy of the participating entities. Following the Weberian tradition, we define \emph{state actors} as formal agents of a recognized government, such as military and police forces, that successfully claim a monopoly on the legitimate use of physical force within a defined territory \cite{Raleigh2023}. Conversely, \emph{non-state actors} are organized, armed groups that operate independently of central government control, ranging from rebel movements seeking national power to communal militias focused on local security or identity \cite{Hu2022}.

\paragraph{Context of Conflict in Nigeria and Cameroon}
Conflict in Africa is characterized by a distinctive landscape that differs fundamentally from patterns observed in the Global North. The continent hosts 46\% of the world's internally displaced persons, with the majority fleeing from violence rather than natural disasters~\cite{IDMC2024}. Unlike conflicts in Western contexts, African conflicts are frequently protracted, multi-causal, rooted in governance failures, and often receive strikingly less international attention and humanitarian funding~\cite{KenyGuyer2018}. These conflicts involve weak state institutions, the proliferation of non-state armed actors, and ethnic fragmentation~\cite{ Mlambo2023}.

Within this context, Nigeria has grappled with the Boko Haram insurgency since 2009. Ranking as one of the world's deadliest terrorist movements at its peak, the group caused 6,644 deaths in 2014 alone~\cite{GlobalTerrorismIndex2015}. The insurgency has resulted in catastrophic humanitarian consequences, including widespread trauma and a 13-percentage-point increase in childhood wasting~\cite{Dunn2018}. The crisis exemplifies how state fragility and the inability to provide basic services create fertile ground for violent extremism~\cite{ProusedeMontclos2014}. Similarly, Cameroon has been engulfed in the Anglophone Crisis since 2016, rooted in the political and social marginalization of English-speaking regions~\cite{Konings1997, Konings2019}. What began as peaceful protests by lawyers and teachers escalated into a secessionist armed conflict after forceful state responses and mass arrests~\cite{Bang2022, Cho2024}. The crisis has resulted in over 6,000 deaths and over half a million civilians displaced~\cite{ICG2022}. Both crises illustrate how West African conflicts are embedded in post-colonial state structures, generating humanitarian catastrophes that receive insufficient international attention despite their scale~\cite{InternationalMonetaryFund2019}.

\paragraph{The Evolution of Computational Systems in Conflict Analysis}
The systematic extraction of political events from text began with rule-based systems that sought to automate monitoring while maintaining interpretability. The Kansas Event Data System (KEDS) pioneered sparse-parsing techniques to identify actor-action-target structures from news reports, achieving plausible agreement with human coders \cite{SchrodtGerner1994}. Subsequent systems like TABARI \cite{Schrodt2001} and PETRARCH integrated the CAMEO (Conflict and Mediation Event Observations) framework designed for non-state actor engagements~\cite{Gerner2002}. However, these approaches are not without validity concerns. As noted by prior research, dictionary-dependent methods in automated text analysis face significant limitations: they require substantial manual curation and are prone to methodological pitfalls, making them unsuitable as substitutes for manual validation~\cite{Grimmer2013}. Hammond et al.\cite{Hammond2014} noted that geographic biases exist in how events are reported in event data and highlighted how they are captured by automated systems. The complex structure of non-Western news reports often pose significant challenges for parsing algorithms, leading to structured errors and reduced validity in less-represented conflict zones.

The transition to neural architectures and LLMs has transformed event coding and analysis while introducing new validity concerns. Domain-specific models like ConfliBERT demonstrated significant improvements over generic BERT in benchmark tasks, particularly when training data was limited ~\cite{Hu2022}. General-purpose LLMs have shown even more dramatic gains; ChatGPT has been found to outperform crowd workers by 25 percentage points on political annotation tasks while achieving high inter-coder agreement at a lower cost~\cite{Gilardi2023}. Zero-shot GPT-4 has similarly outperformed expert human classifiers for political identification across multiple countries~\cite{Tornberg2024}. These capabilities enable deployment in humanitarian response at unprecedented scales, such as the Violence Early-Warning System (ViEWS)~\cite{Hegre2019, Hegre2021}.

\paragraph{Strategic Risks of Employing LLMs in International Policy}
Disparities in LLM performance, particularly the emergence of \textit{state-actor neutrality bias}, constitute more than a technical limitation. As interest grows in applying such systems to conflict monitoring contexts \cite{Hegre2019, Sanchez2025}, the systematic normative distortions we document raise important questions about their reliability for human security applications. The operational impact of misclassification is amplified by \textit{automation bias}, whereby human analysts disproportionately defer to algorithmic outputs even when they conflict with local reporting or contextual knowledge \cite{ISQ2024, AlonBarkat2023}. In settings where models implicitly encode a posterior that treats state actors as less violent or more legitimate than non-state groups, this deference could contribute to systematic under-reporting of state-led violence while over-attributing violence to non-state actors, a distortion with direct implications for resource allocation and protection decisions in conflict-affected populations.

These risks are further compounded by narrative distortion in politically fragile environments such as Cameroon’s Anglophone regions or Nigeria’s Middle Belt, where the boundary between peaceful dissent and armed conflict is highly contested \cite{Bang2022, Cho2024}. Misclassification that attributes violence to non-state actors in the absence of empirical evidence risks lending apparent objectivity to coercive state responses, a concern of particular consequence in contexts where peaceful dissent and armed conflict are not clearly demarcated \cite{Bang2022, Cho2024}.

\section{Methodology}
\label{sec:methodology}

\subsection{Data and Task Definition}
We utilize a consolidated dataset from ACLED covering Nigeria and Cameroon from 2020--2024. Each record consists of structured metadata and unstructured textual descriptions (``notes''). The prediction task is a multi-class classification over six canonical categories: \textit{Violence against civilians (V)}, \textit{Battles (B)}, \textit{Explosions (E)}, \textit{Protests (P)}, \textit{Riots (R)}, and \textit{Strategic developments (S)}. This label space is fixed across all models, prompting strategies, and evaluation stages to ensure consistency and comparability. To also avoid confounding effects from sample variation, all models operate on identical, shared subsets of events for each country.

\subsection{Pre-processing and Actor Normalization}
We apply a standardized pipeline to harmonize heterogeneous fields. Records are de-duplicated using unique ACLED identifiers. A central component is actor normalization, which maps raw mentions to canonical groups using country-aware heuristics (e.g., ``Military'' $\rightarrow$ State Actor; ``Rebel Group'' $\rightarrow$ Non-State Actor). This abstraction preserves the semantic role of actors while enabling group-based fairness analysis. We sample 1000 event data points per country with equal proportions of state and non-state actors. For prompt-based evaluations, we construct curated datasets emphasizing data quality and class balance. Candidate examples are scored based on description length and the presence of identifiable actors. We also carry out stratified sampling to ensure balanced representation across all event classes.

\subsection{Modeling Framework} We evaluate two paradigms broadly: domain-specific architectures and open-weight general purpose LLMs. The domain-specific models include AfroConfliBERT, a derivative of ConfliBERT \cite{Hu2022}, and AfroConfliLLAMA, obtained from Llama via LoRA. The general purpose LLMs are prompt-based LLMs namely Llama, Gemma, Olmo, and Mistral, which are treated as "off-the-shelf" black-box systems. We study small- to mid-sized open-weight models because they allow for faster experimentation and require less inference costs compared to frontier models. All distinctions of \textit{domain-specific} architectures and \textit{general purpose} or \textit{open-weight} or \textit{vanilla} models in this paper stem from this dichotomy.

\section{Experimental Design}
\label{sec:experimental_design}

\paragraph{System Configuration and Reproducibility}
All experiments use fixed random seeds to ensure reproducibility. We employ bootstrapping to estimate confidence intervals and permutation testing to assess the significance of group disparities and counterfactual effects. For continuous measures, both parametric and non-parametric tests are reported to ensure robustness to distributional assumptions. Experiments are executed on a single high-memory A100 GPU for prompt-based inference. To foster open science and ensure full methodological transparency, we disclose our generation parameters, prompting strategies and few-shot selections in section \ref{appendix:experimental_reproducibility} of the appendix.

\paragraph{Calibration and Uncertainty} To align predicted confidence scores with empirical accuracy, we apply post-hoc calibration using \textbf{isotonic regression} to learn a monotonic mapping and \textbf{temperature scaling} to rescale logits via a single parameter. We assess calibration quality using Brier scores (see Appendix~\ref{appendix:calibration_metrics}) and analyze coverage--accuracy trade-offs to determine how confidence thresholds influence selective prediction.

\paragraph{Event Ambiguity Scoring}
\label{subsec:event_ambiguity_scoring}
In order to ensure our studies of inter-model disagreements and perturbations are void of task-inherent uncertainty, we develop a novel event ambiguity scoring (EAS) technique, based on previous studies of aleatoric and epistemic uncertainty ~\cite{kendall2017uncertainties}. We expound on its details in section \ref{appendix:event_ambiguity_scoring} of the appendix. All our counterfactual and error trace analyses are carried out on low-ambiguity events to minimize the effect of aleatoric uncertainty.

\paragraph{Fairness and Counterfactual Analysis}
\label{subsec:fairness_and_counterfactual_analysis}
To assess the socio-technical impact of model deployment in conflict zones, we conduct a two-tiered evaluation focused on both group-level disparities and individual-level causal sensitivity:
\begin{itemize}
    \item \textbf{Tier 1: Group Fairness.} We evaluate performance disparities between \textit{State} and \textit{Non-State} actor events. We quantify these using \textit{Statistical Parity Difference} (SPD) and \textit{Equalized Odds}, the latter of which measures differences in True Positive Rates (TPR) and False Positive Rates (FPR) across groups to identify coherent patterns of  misclassification bias.
    \item \textbf{Tier 2: Counterfactual Sensitivity.} We assess identifying potential causal dependencies in model reasoning by applying controlled perturbations to event descriptions. These modifications involve the targeted removal or insertion of actor references (e.g., changing ``Police'' to ``Unidentified Armed Group'') while preserving the underlying event mechanics.
\end{itemize}

\noindent
Disparities and sensitivity metrics are validated using 95\% bootstrap confidence intervals and non-parametric permutation-based significance tests. Throughout the experimental process, perturbation quality is strictly monitored via manual spot-checks to ensure linguistic plausibility and to prevent the introduction of artefactual cues that could lead to trivial classification shifts.

\subsection{Aggregation and Analysis Pipeline}
The workflow follows a multi-stage process: \textit{Inference} $\rightarrow$ \textit{Calibration} $\rightarrow$ \textit{Metric Computation} $\rightarrow$ \textit{Fairness, Counterfactual and Error Trace Analysis}. Intermediate artifacts are stored in a structured hierarchy organized by country and modeling strategy. We use merge-based procedures rather than overwriting artifacts, enabling cumulative and retrospective analysis.

\section{Results}
\label{sec:results}

\subsection{Classification metrics}

Tables \ref{tab:model_perf_combined} shows a clear geography-based performance gap on macro classification metrics, with all models achieving substantially higher accuracy on Nigeria than on Cameroon. AfroConfliBERT and AfroConfliLLAMA noticeably consistently outperform others across both regions. Figure \ref{fig:f1_comparison_countries_zero_shot} illustrates the F1 scores for classification labels across the five models in the study. Again, AfroConfliBERT and AfroConfliLLAMA lead with robust $F_1$ scores (often $>0.9$).

\begin{table}[t]
\centering
\small
\caption{Model performance on conflict classification for Cameroon and Nigeria. Metrics shown are Accuracy (Acc), Macro Precision (P), and Macro Recall (R).}
\label{tab:model_perf_combined}
\begin{tabular}{l c ccc ccc}
\toprule
& & \multicolumn{3}{c}{\textbf{Cameroon}} & \multicolumn{3}{c}{\textbf{Nigeria}} \\
\cmidrule(lr){3-5} \cmidrule(lr){6-8}
\textbf{Model} & \textbf{N} & \textbf{Acc} & \textbf{P} & \textbf{R} & \textbf{Acc} & \textbf{P} & \textbf{R} \\
\midrule
AfroConfliBERT & 1000 & 0.943 & 0.935 & 0.942 & 0.972 & 0.964 & 0.978 \\
Llama 3.2 3B & 1000 & 0.688 & 0.641 & 0.654 & 0.785 & 0.797 & 0.678 \\
Gemma 3 4B   & 1000 & 0.618 & 0.68 & 0.672 & 0.654 & 0.73 & 0.591 \\
Olmo2 7B     & 1000 & 0.662 & 0.499 & 0.671 & 0.76 & 0.741 & 0.583 \\
Mistral 7B   & 1000 & 0.699 & 0.539 & 0.752 & 0.793 & 0.735 & 0.668 \\
AfroConfliLLAMA & 1000 & 0.964  & 0.938 & 0.96  & 0.97  & 0.943 & 0.983 \\
\bottomrule
\end{tabular}
\end{table}

\subsection{Comparative Analysis of Confusion Matrices}

As observed in figures \ref{fig:confusion_matrices_nigeria} and \ref{fig:confusion_matrices_cameroon}, label \textit{P} (protests) consistently emerges as a distinct linguistic anchor point across both datasets with near-perfect diagonal accuracy. While AfroConfliBERT and AfroConfliLLAMA demonstrate superior zero-shot robustness, regional nuances create specific failure modes: the Nigeria dataset exhibits significant semantic scattering for labels B and R, whereas the Cameroon dataset is characterized by a pervasive S-to-V confusion.

\subsection{Fairness Evaluation}
We evaluate disparities across state and non-state actors. Specifically, we measure Statistical Parity Difference (SPD) to assess group-level selection bias and Equalized Odds (TPR/FPR disparities) to assess group-level error bias. Our results for Cameroon and Nigeria are presented in Table \ref{tab:fairness_results_combined}.

\begin{table}[t]
\centering
\small
\caption{Fairness metrics for label V (Violence) across actor groups in Cameroon and Nigeria. SPD measures selection bias; $\Delta$TPR and $\Delta$FPR denote error disparities (State vs.\ Non-State reference). $P$-values are from permutation tests ($N=1000$). Bold $P$-values indicate significant disparity ($p < 0.05$).}
\label{tab:fairness_results_combined}
\begin{tabular}{l ccccc ccccc}
\toprule
& \multicolumn{5}{c}{\textbf{Cameroon}} & \multicolumn{5}{c}{\textbf{Nigeria}} \\
\cmidrule(lr){2-6} \cmidrule(lr){7-11}
\textbf{Model}
& \textbf{SPD} & $\Delta$TPR & $P_{TPR}$ & $\Delta$FPR & $P_{FPR}$
& \textbf{SPD} & $\Delta$TPR & $P_{TPR}$ & $\Delta$FPR & $P_{FPR}$ \\
\midrule
AfroConfliBERT
& $-0.058$ & $-0.010$ & $0.639$ & $+0.084$ & $\mathbf{0.000}$
& $-0.307$ & $-0.040$ & $0.241$ & $-0.006$ & $0.523$ \\
Gemma
& $-0.040$ & $-0.007$ & $0.389$ & $+0.029$ & $0.483$
& $-0.182$ & $-0.012$ & $0.402$ & $+0.026$ & $0.463$ \\
Llama
& $-0.112$ & $-0.033$ & $\mathbf{0.014}$ & $-0.048$ & $0.243$
& $-0.364$ & $+0.000$ & $1.000$ & $-0.177$ & $\mathbf{0.000}$ \\
Mistral
& $-0.164$ & $+0.029$ & $0.446$ & $-0.145$ & $\mathbf{0.000}$
& $-0.323$ & $-0.225$ & $\mathbf{0.000}$ & $-0.053$ & $\mathbf{0.001}$ \\
Olmo
& $-0.102$ & $+0.089$ & $\mathbf{0.044}$ & $-0.083$ & $\mathbf{0.001}$
& $-0.327$ & $-0.125$ & $\mathbf{0.016}$ & $-0.073$ & $\mathbf{0.000}$ \\
AfroConfliLLAMA
& $-0.116$ & $-0.038$ & $0.070$ & $+0.013$ & $0.364$
& $-0.317$ & $-0.066$ & $\mathbf{0.026}$ & $-0.008$ & $0.307$ \\
\bottomrule
\end{tabular}
\end{table}

\subsubsection{Analysis of Actor-Based Disparities in Cameroon}
Our analysis of the Cameroonian dataset reveals regular disparities that suggest models carry latent posteriors regarding the roles of state versus non-state entities.

\begin{itemize}
\item \textbf{Selection Bias (SPD)}: All evaluated models exhibit negative Statistical Parity Difference. Mistral shows the most disparity (SPD $= -0.164$), indicating it is over 16\% less likely to assign a violence label to a state-actor event compared to a non state-actor event. Since the 95\% Confidence Intervals do not cross zero for any LLM, we can conclude that both fine-tuned and general LLMs possess a statistically significant \textbf{state-actor neutrality} bias.

\item \textbf{Sensitivity Disparity ($\Delta$TPR)}: Llama exhibits a statistically significant negative disparity in True Positive Rates ($p = 0.014$). This suggests a blind spot where actual state-led violence is more likely to be misclassified as non-violent. Interestingly, Olmo demonstrates a significant positive disparity ($\Delta$TPR $= +0.089$, $p = 0.044$), indicating it is uniquely more sensitive to identifying state violence than non-state violence in this context.

\item \textbf{False Attribution ($\Delta$FPR)}: Several models demonstrate critical failures in False Positive parity. Mistral and Olmo exhibit significant negative disparities ($p \le 0.001$), meaning they are much more likely to falsely attribute violence to non-state actors. Conversely, the specialized AfroConfliBERT model exhibits a significant positive disparity ($\Delta$FPR $= +0.084$, $p < 0.001$), making it more prone to falsely attributing violence to state actors.
\end{itemize}

\subsubsection{Analysis of Actor-Based Disparities in Nigeria}
The results indicate that the Nigerian context exacerbates the state-actor neutrality bias observed in Cameroon.

\begin{itemize}
\item \textbf{Selection Bias (SPD)}: The Statistical Parity Difference in Nigeria is significantly more pronounced. Llama displays a staggering SPD $= -0.364$, indicating that the model is over 36\% less likely to predict a violent label when the subject is a state actor. Even fine-tuned models like AfroConfliLLAMA ($-0.317$) and AfroConfliBERT ($-0.307$) exhibit severe selection bias, suggesting a deeply embedded posterior that strongly associates "state" with non-violent or legitimate status.

\item \textbf{Sensitivity Disparity ($\Delta$TPR)}: Mistral, Olmo, and AfroConfliLLAMA demonstrate highly significant and severe disparities in True Positive Rates ($p < 0.05$). Specifically, Mistral is 22.5\% less likely to correctly identify an actual violent event if it is perpetrated by a state actor compared to a non-state actor. This represents a functional "blindness" to state-led violence.

\item \textbf{False Attribution ($\Delta$FPR)}: While Gemma and the ACLED fine-tuned models maintain a relatively equitable false positive distribution, Llama exhibits a severe False Positive disparity ($p < 0.001$). It is 17.7\% more likely to falsely attribute violence to non-state actors than to state actors, effectively criminalizing non-state actor behavior even in non-violent scenarios (e.g., peaceful protests). Mistral and Olmo mirror this trend with statistically significant negative disparities.
\end{itemize}

\noindent
Our findings imply that for automated conflict monitoring, reliance on these models without bias-correction will result in data that downplays state-actor involvement in violent events while inflating the violent profile of non-state actors.

\subsection{Counterfactual Analysis}

\subsubsection{Lexical Sensitivity}

To evaluate the stability of model reasoning, we conduct a word-level perturbation analysis. By inserting or substituting specific lexical markers, we measure the \textit{Flip Rate}, the rate at which inserting the lexical marker changes the predicted label, as well as the \textit{Confidence Shift} ($\Delta\phi$) relative to the baseline. The results, summarized in Tables \ref{tab:word_level_perturbation_sensitivity_cmr} and \ref{tab:word_level_perturbation_sensitivity_nga}, reveal that the models' classification of conflict events is not merely a function of described actions, but rather a function of legitimizing framing and epistemic signaling.

\begin{itemize}
\item \textbf{High Vulnerability to Explicit Delegitimization}:
The models exhibit their most drastic instabilities when presented with language that explicitly delegitimizes an actor's conduct. In the Cameroon dataset, inserting the phrase \textit{violating human rights} generates the highest observed flip rate at 66.7\%, yielding a massive and statistically significant confidence shift ($\Delta\phi = +54.44, p = 0.005$) against a baseline noise of only 12.2\%. This trend is strongly reinforced in the Nigeria dataset, where the delegitimizing phrases \textit{unprovoked} and \textit{using excessive force} drive the highest sample-size disruptions (N=114 each). These perturbations result in flip rates of 21.1\% and 19.3\%, respectively, both achieving high statistical significance ($p < 0.001$) against a baseline of 4.4\%. This indicates that the models heavily index on overt moral or legal judgments inserted into the text, allowing subjective framing to override the underlying factual taxonomy of the event.

\item \textbf{Sensitivity to Action Substitutions and Intensity Markers}:
The models demonstrate significant brittleness to the specific verbs and adverbs used to describe kinetic events. In the Cameroon test set, substituting a standard action verb with \textit{engaged} causes a 50.0\% label flip rate ($\Delta\phi = +37.78, p = 0.039$), while adding the intensity marker \textit{brutally} shifts the prediction in 26.2\% of cases ($\Delta\phi = +13.97, p = 0.045$). A parallel effect occurs in the Nigeria data, where the action substitution \textit{executed} ($\Delta\phi = +10.43, p = 0.028$) and the negation marker \textit{killed} ($\Delta\phi = +12.28, p = 0.014$) both generate statistically significant deviations from the baseline. The models' decision boundaries are therefore highly permeable to stylistic vocabulary choices that imply either formal military parity (\textit{engaged}) or unilateral severity (\textit{executed, brutally}).

\item \textbf{Contextual and Regional Divergence in Lexical Weights}:
The data further reveals that the models do not apply these lexical sensitivities uniformly across all contexts. While delegitimizing terms like \textit{unprovoked} and \textit{using excessive force} are highly disruptive and statistically significant in the Nigerian context ($p < 0.001$), the exact same structural perturbations fail to achieve statistical significance in the Cameroonian context ($p = 0.373$ and $p = 0.579$, respectively). Conversely, action substitutions that trigger major, significant shifts in Cameroon (such as \textit{engaged}) do not appear as top-tier significant drivers in Nigeria. This discrepancy suggests that the models maintain region-specific lexical priors; the epistemic weight of a specific framing device depends heavily on the geographic or sovereign context in which the event is situated.
\end{itemize}

\noindent
Our lexical sensitivity profiling indicates that these models do not possess a robust, act-based understanding of conflict. Instead, they rely on a fragile network of linguistic cues that prioritize \textit{how} an event is framed over \textit{what} actually occurred. Because statistically significant label flips can be reliably engineered simply by inserting delegitimizing phrases, altering verbs, or adding intensity adverbs, all without changing the core factual sequence of the event, the models operate less as objective event classifiers and more as semantic framing evaluators.

\begin{table}[t]
\centering
\caption{Statistical Details for Word-Level Perturbation Sensitivity on Cameroonian events.}
\label{tab:word_level_perturbation_sensitivity_cmr}
\scriptsize
\begin{tabular}{@{}l l r r r r r r@{}}
\toprule
\textbf{Word / Phrase} & \textbf{Perturbation Type} & \textbf{N} & \textbf{Flip \%} & \textbf{95\% CI} & \textbf{$\Delta\phi$} & \textbf{$d$} & \textbf{$p$} \\
\midrule
violating human rights & Delegitimation & 6 & 66.7 & (22.3--95.7) & $+54.44$ & 1.20 & 0.005 \\
engaged & Action substitution & 6 & 50.0 & (11.8--88.2) & $+37.78$ & 0.86 & 0.039 \\
brutally & Intensity & 42 & 26.2 & (13.9--42.0) & $+13.97$ & 0.36 & 0.045 \\
Government officials stated that & Provenance & 6 & 33.3 & (4.3--77.7) & $+21.11$ & 0.52 & 0.186 \\
fought & Action substitution & 6 & 33.3 & (4.3--77.7) & $+21.11$ & 0.52 & 0.186 \\
confronted & Action substitution & 6 & 33.3 & (4.3--77.7) & $+21.11$ & 0.52 & 0.186 \\
murdered & Action substitution & 30 & 20.0 & (7.7--38.6) & $+7.78$ & 0.21 & 0.364 \\
killed & Negation & 30 & 20.0 & (7.7--38.6) & $+7.78$ & 0.21 & 0.364 \\
violently & Intensity & 30 & 20.0 & (7.7--38.6) & $+7.78$ & 0.21 & 0.364 \\
unprovoked & Delegitimation & 114 & 16.7 & (10.3--24.8) & $+4.44$ & 0.13 & 0.373 \\
Human rights organizations documented that & Provenance & 6 & 16.7 & (0.4--64.1) & $+4.44$ & 0.13 & 0.562 \\
without justification & Delegitimation & 6 & 16.7 & (0.4--64.1) & $+4.44$ & 0.13 & 0.562 \\
attacked & Action substitution & 6 & 16.7 & (0.4--64.1) & $+4.44$ & 0.13 & 0.562 \\
struck & Action substitution & 6 & 16.7 & (0.4--64.1) & $+4.44$ & 0.13 & 0.562 \\
beat & Negation & 6 & 16.7 & (0.4--64.1) & $+4.44$ & 0.13 & 0.562 \\
using excessive force & Delegitimation & 114 & 14.9 & (8.9--22.8) & $+2.69$ & 0.08 & 0.579 \\
\midrule
\textit{Neutral Control} & \textit{Baseline Noise} & 90 & 12.2 & (6.3--20.8) & -- & -- & -- \\
\bottomrule
\end{tabular}
\vspace{0.5em}
\\[0.5em]
\small

\vspace{1mm}
\footnotesize
\raggedright
\textit{Note:} Flip \% denotes the rate at which inserting the word or phrase changes the predicted label.
95\% Confidence Intervals are calculated safely using the Clopper-Pearson exact method for binomial proportions.
$\Delta\phi$ denotes shift relative to the baseline in percentage points.
$d$ is Cohen’s effect size.
$p$-values use Pearson's $\chi^2$ (or Fisher's Exact test dynamically if any expected cell frequency $< 5$).
Statistics are aggregated across all models. \textit{N} denotes the number of perturbation instances across all models and events. Substitutions are triggered only when the source term appears in the original text. Entity-specific perturbations (e.g. actor) occur less frequently than structural perturbations (e.g. decontextualization) due to within-sample variations resulting in lower $N$ values and wider confidence intervals.
\end{table}

\subsubsection{Model Vulnerability to Semantic Shifts}

While the aggregate lexical sensitivity identifies universal triggers, a model-stratified analysis is shown in Tables \ref{tab:vulnerability_stats_cmr} and \ref{tab:vulnerability_stats_nga}. Based on the empirical data across both national contexts, three core deductions characterize model-specific vulnerabilities:

\begin{itemize}
    \item \textbf{Resilience of Domain-Adapted Architectures}:
    The data demonstrates that domain-specific models are largely impervious to superficial semantic perturbations. AfroConfliBERT exhibits near-absolute stability, recording a 0.0\% flip rate across all eight perturbation categories in the Cameroon test set, and registering negligible shifts (max 7.9\% for delegitimation) in the Nigeria set. AfroConfliLLAMA similarly maintains strict boundary adherence, with flip rates rarely exceeding 11\% across any category in either dataset. For these models, confidence shifts ($\Delta\phi$) remain functionally flat, and no perturbations achieve statistical significance against the baseline.
    \item \textbf{Susceptibility of Open-Weight Models to Delegitimization}:
    In contrast, open-weight models demonstrate severe volatility when exposed to subjective framing, particularly delegitimization. In the Nigeria test set, delegitimization framing is the only perturbation type to yield statistically significant vulnerabilities, driving a 34.2\% flip rate in Olmo 2 7B (p = 0.022, Cohen’s h = 0.79) and a 26.3\% flip rate in Mistral (p = 0.022, Cohen's h = 1.08). This trend mirrors the Cameroon data, where Olmo and Mistral flipped at rates of 37.5\% and 32.5\%, respectively, when presented with delegitimizing language. This confirms that general-purpose models systematically lack the epistemic defense mechanisms required to separate factual conduct from rhetorical condemnation.
    \item \textbf{Volatility to Intensity and Action Modifiers}:
    Beyond moral framing, open-weight models also exhibit distinct brittleness regarding the intensity and mechanical description of actions. In Cameroon, modifying the intensity of an action caused Llama to alter its classification 33.3\% of the time, while Olmo flipped 29.2\% of the time. Similarly, action substitutions triggered high instability in Olmo (33.3\%) and Mistral (23.8\%).
\end{itemize}

\begin{table}[htbp]
\centering
\caption{Model vulnerability to semantic perturbations for Cameroon. Flip \% denotes the rate at which a perturbation changes the predicted label. $\Delta\phi$ is the mean confidence shift over the baseline. $p$ values reflect comparison of perturbation flip rates against neutral controls (Fisher's exact test). $h$ is Cohen's $h$ characterizing the overall effect size of those proportional differences.}
\label{tab:vulnerability_stats_cmr}
\scriptsize
\begin{tabular}{@{}l l r c r r l@{}}
\toprule
\textbf{Perturbation Type} &
\textbf{Model} &
\textbf{Flip \%} &
\textbf{95\% CI} &
\textbf{$\Delta\phi$} &
\textbf{$h$} &
\textbf{$p$} \\
\midrule
\multirow{6}{*}{Legitimation framing}
 & AfroConfliBERT      &   0.0 & (0.0--16.8)   & $-0.16$ &  0.00 & 1.000 \\
 & Gemma     &   5.0 & (0.1--24.9)   & $-0.50$ & -0.07 & 1.000 \\
 & Llama    &  20.0 & (5.7--43.7)   & $+1.83$ & -0.16 & 0.700 \\
 & Mistral     &  15.0 & (3.2--37.9)   & $+1.17$ &  0.05 & 1.000 \\
 & Olmo       &  10.0 & (1.2--31.7)   & $-1.67$ & -0.10 & 1.000 \\
& AfroConfliLLAMA       &  10.0 & (1.2--31.7)   & $+0.00$ & -0.10 & 1.000 \\
\addlinespace
\multirow{6}{*}{Delegitimation framing}
 & AfroConfliBERT      &   0.0 & (0.0--8.8)    & $-0.04$ &  0.00 & 1.000 \\
 & Gemma      &  10.0 & (2.8--23.7)   & $+1.25$ &  0.12 & 1.000 \\
 & Llama   &  17.5 & (7.3--32.8)   & $-5.42$ & -0.22 & 0.468 \\
 & Mistral      &  32.5 & (18.6--49.1)  & $-2.33$ &  0.47 & 0.192 \\
 & Olmo      &  37.5 & (22.7--54.2)  & $-0.42$ &  0.57 & 0.109 \\
 & AfroConfliLLAMA       &   5.0 & (0.6--16.9)   & $+0.00$ & -0.30 & 0.298 \\
\addlinespace
\multirow{6}{*}{Provenance addition}
 & AfroConfliBERT      &   0.0 & (0.0--6.0)    & $-0.01$ &  0.00 & 1.000 \\
 & Gemma     &   6.7 & (1.8--16.2)   & $+1.67$ &  0.00 & 1.000 \\
 & Llama   &  10.0 & (3.8--20.5)   & $-1.50$ & -0.44 & 0.105 \\
 & Mistral     &  16.7 & (8.3--28.5)   & $-3.58$ &  0.09 & 1.000 \\
 & Olmo      &  36.7 & (24.6--50.1)  & $-1.17$ &  0.55 & 0.123 \\
 & AfroConfliLLAMA       &   5.0 & (1.0--13.9)   & $+0.00$ & -0.30 & 0.260 \\
\addlinespace
\multirow{6}{*}{Intensity modification}
 & AfroConfliBERT      &   0.0 & (0.0--14.2)   & $-0.01$ &  0.00 & 1.000 \\
 & Gemma      &   0.0 & (0.0--14.2)   & $+3.33$ & -0.52 & 0.385 \\
 & Llama   &  33.3 & (15.6--55.3)  & $+1.75$ &  0.15 & 0.734 \\
 & Mistral     &  25.0 & (9.8--46.7)   & $+1.33$ &  0.30 & 0.450 \\
 & Olmo       &  29.2 & (12.6--51.1)  & $+2.25$ &  0.39 & 0.437 \\
 & AfroConfliLLAMA       &  12.5 & (2.7--32.4)   & $+0.00$ & -0.02 & 1.000 \\
\addlinespace
\multirow{6}{*}{Negation}
 & AfroConfliBERT      &   0.0 & (0.0--24.7)   & $-0.01$ &  0.00 & 1.000 \\
 & Gemma      &   7.7 & (0.2--36.0)   & $-0.77$ &  0.04 & 1.000 \\
 & Llama   &  23.1 & (5.0--53.8)   & $+0.56$ & -0.08 & 1.000 \\
 & Mistral     &  23.1 & (5.0--53.8)   & $+1.59$ &  0.25 & 0.639 \\
 & Olmo       &  30.8 & (9.1--61.4)   & $-1.05$ &  0.43 & 0.372 \\
 & AfroConfliLLAMA       &   7.7 & (0.2--36.0)   & $+0.00$ & -0.19 & 1.000 \\
\addlinespace
\multirow{6}{*}{Actor substitution}
 & AfroConfliBERT      &   0.0 & (0.0--7.9)    & $+0.00$ &  0.00 & 1.000 \\
 & Gemma      &  11.1 & (3.7--24.1)   & $+0.44$ &  0.16 & 1.000 \\
 & Llama   &  11.1 & (3.7--24.1)   & $+1.56$ & -0.41 & 0.208 \\
 & Mistral     &  13.3 & (5.1--26.8)   & $-0.33$ &  0.00 & 1.000 \\
 & Olmo       &  11.1 & (3.7--24.1)   & $-0.33$ & -0.07 & 1.000 \\
 & AfroConfliLLAMA       &   6.7 & (1.4--18.3)   & $+0.00$ & -0.23 & 0.591 \\
\addlinespace
\multirow{6}{*}{Action substitution}
 & AfroConfliBERT      &   0.0 & (0.0--16.1)   & $-0.06$ &  0.00 & 1.000 \\
 & Gemma      &  14.3 & (3.0--36.3)   & $+0.95$ &  0.25 & 0.626 \\
 & Llama   &   9.5 & (1.2--30.4)   & $+2.29$ & -0.46 & 0.210 \\
 & Mistral     &  23.8 & (8.2--47.2)   & $+1.10$ &  0.27 & 0.674 \\
 & Olmo       &  33.3 & (14.6--57.0)  & $+0.76$ &  0.48 & 0.252 \\
 & AfroConfliLLAMA       &  14.3 & (3.0--36.3)   & $+0.00$ &  0.03 & 1.000 \\
\addlinespace
\multirow{6}{*}{Decontextualization}
 & AfroConfliBERT      &   0.0 & (0.0--10.3)   & $-0.14$ &  0.00 & 1.000 \\
 & Gemma      &   0.0 & (0.0--10.3)   & $+0.29$ & -0.52 & 0.306 \\
 & Llama   &  14.7 & (5.0--31.1)   & $+1.04$ & -0.30 & 0.427 \\
 & Mistral     &   0.0 & (0.0--10.3)   & $+0.99$ & -0.75 & 0.089 \\
 & Olmo       &  11.8 & (3.3--27.5)   & $-0.08$ & -0.05 & 1.000 \\
 & AfroConfliLLAMA       &   0.0 & (0.0--10.3)   & $+0.00$ & -0.75 & 0.089 \\
\addlinespace
\midrule
\multicolumn{7}{l}{\textbf{Neutral Controls}} \\
 & AfroConfliBERT      &   0.0 & (0.0--21.8) & -0.00 & -- & -- \\
 & Gemma      &   6.7 & (0.2--31.9) & +0.00 & -- & -- \\
 & Llama   &  26.7 & (7.8--55.1) & -1.33 & -- & -- \\
 & Mistral     &  13.3 & (1.7--40.5) & +0.33 & -- & -- \\
 & Olmo       &  13.3 & (1.7--40.5) & +0.67 & -- & -- \\
 & AfroConfliLLAMA       &  13.3 & (1.7--40.5) & +0.00 & -- & -- \\
\bottomrule
\end{tabular}
\end{table}

\subsubsection{Sensitivity Clusters}

Sensitivity clusters group examples that behave similarly when we perturb inputs (e.g., swap/remove actors, change phrasing, or apply counterfactual edits) and summarize how fragile each group is to those changes. These clusters therefore make model brittleness modes easy to spot, prioritize for annotation/augmentation, and monitor for fairness or robustness interventions. Figure \ref{fig:sensitivity_clusters} displays a few sensitivity clusters obtained from the zero-shot runs of AfroConfliBERT, Olmo, Gemma and Mistral on the Cameroon dataset.

\subsection{Legitimization Bias and Normative Directionality}

To evaluate whether models exhibit systemic bias in the classification of political violence, we analyze the directionality of errors between battles (\textit{B}) 
and violence against civilians (\textit{V}). We define the Legitimization Bias 
Difference ($\Delta_{\mathrm{LB}}$) as the difference between the False 
Illegitimation rate ($\varepsilon_{\mathrm{FI}}$) and the False Legitimation 
rate ($\varepsilon_{\mathrm{FL}}$) as shown in Tables \ref{tab:legitimization_bias_nga} and \ref{tab:legitimization_bias_cmr}. Our results reveal a bifurcated divergence in normative directionality between general-purpose models and domain-specialized architectures.

\begin{itemize}
\item \textbf{False Illegitimation Bias in General Purpose LLMs}: Across both the Nigeria and Cameroon test sets, the open-weight models, most notably Gemma and Llama, exhibit an asymmetric bias skewed toward False Illegitimation ($\Delta_{\mathrm{LB}} \gg 0$). For Gemma, the \textit{FI} rate reaches 11.25\% in Nigeria and scales to 18.29\% in Cameroon, while its \textit{FL} rate remains strictly at 0\% across both datasets. This asymmetry is highly statistically significant ($p < 10^{-8}$ for Nigeria, $p < 10^{-17}$ for Cameroon). Llama demonstrates a similarly entrenched bias, misclassifying symmetric engagements between armed groups (Battles) as unilateral attacks on non-combatants at moderate rates of 6.11\% and 10.32\%, respectively. Even in models where the bias is less pronounced, such as Mistral and Olmo, a shift toward False Illegitimation becomes statistically significant in the Cameroon set ($p=0.022$ and $p=0.051$, respectively).

\item \textbf{Parity in Domain-Adapted Architectures}: In contrast to the vanilla LLMs, AfroConfliBERT and AfroConfliLLAMA achieve directional neutrality. These models maintain $\Delta_{\mathrm{LB}}$ values close to zero and their error distributions remain largely symmetric. AfroConfliLLAMA showed a slight negative bias in both sets ($-0.13$ pp in Nigeria; $-0.54$ pp in Cameroon), indicating a negligible shift toward False Legitimation. However, these results were statistically non-significant ($p=0.762$ and $p=0.354$). AfroConfliBERT recorded a $\Delta_{\mathrm{LB}}$ of $+0.98$ pp in Nigeria, the $p$-value ($0.100$) fails to meet standard significance thresholds. In Cameroon, its $\Delta_{\mathrm{LB}}$ dropped further to $+0.31$ pp ($p=0.536$).
\end{itemize}

\begin{table}[t]
\centering
\caption{Legitimization error rates by model on the Cameroon test set
($|V| = 362$ true violence events, $|B| = 339$ true battle events).
$n_{\mathrm{FL}}$ denotes V$\rightarrow$B errors and
$n_{\mathrm{FI}}$ denotes B$\rightarrow$V errors.
Error rates $\varepsilon$ are reported as percentages with 95\% Wilson score confidence intervals.
$\Delta_{\mathrm{LB}} = \varepsilon_{\mathrm{FI}} - \varepsilon_{\mathrm{FL}}$ (percentage points).
$p$ denotes the two-tailed two-proportion $z$-test.}
\label{tab:legitimization_bias_cmr}
\small
\setlength{\tabcolsep}{6pt}
\renewcommand{\arraystretch}{1.15}

\begin{tabular}{l r r l r r l r r}
\toprule
Model
& $n_{\mathrm{FL}}$
& $\varepsilon_{\mathrm{FL}}$ (\%)
& 95\% CI$_{\mathrm{FL}}$
& $n_{\mathrm{FI}}$
& $\varepsilon_{\mathrm{FI}}$ (\%)
& 95\% CI$_{\mathrm{FI}}$
& $\Delta_{\mathrm{LB}}$ (pp)
& $p$ \\
\midrule
AfroConfliBERT
& 1 & 0.28 & [0.05, 1.54]
& 2 & 0.59 & [0.16, 2.12]
& $+0.31$ & 0.536 \\

Gemma
& 0 & 0.00 & [0.00, 1.03]
& 62 & 18.29 & [14.59, 22.68]
& $+18.29$ & $1.8 \times 10^{-18}$ \\

Llama
& 0 & 0.00 & [0.00, 1.03]
& 35 & 10.32 & [7.56, 13.94]
& $+10.32$ & $1.2 \times 10^{-10}$ \\

Mistral
& 6 & 1.66 & [0.76, 3.56]
& 16 & 4.72 & [2.92, 7.53]
& $+3.06$ & 0.022 \\

Olmo
& 6 & 1.66 & [0.76, 3.56]
& 14 & 4.13 & [2.48, 6.81]
& $+2.47$ & 0.051 \\

AfroConfliLLAMA
& 3 & 0.83 & [0.28, 2.40]
& 1 & 0.29 & [0.05, 1.65]
& $-0.54$ & 0.354 \\
\bottomrule
\end{tabular}

\vspace{1mm}
\footnotesize
\raggedright
\textit{Statistical notes.}
$\epsilon_{\mathrm{FL}} = n_{\mathrm{FL}} / |V|$ and
$\epsilon_{\mathrm{FI}} = n_{\mathrm{FI}} / |B|$.
Confidence intervals are Wilson score intervals ($z=1.96$).
$p$ values are from a two-sided two-proportion $z$-test with pooled variance;
for cells with small counts or zeros, these values should be interpreted as
approximations.
\end{table}

\subsection{Effect of In-Context Learning on Legitimization Bias}

A longitudinal comparison across zero-shot, 3-shot, and 5-shot prompting reveals that ICL functions as an inconsistent mechanism for normative calibration. For Llama, ICL serves as an effective stabilizer, moving the model from a slight illegitimation bias in the 3-shot setting ($\Delta_{\mathrm{LB}} = +1.46$) to near-perfect statistical neutrality in the 5-shot condition ($\Delta_{\mathrm{LB}} = -0.05, p = 0.970$). However, for larger models, this calibration is remarkably fragile. As the number of shots increases, models that appeared relatively stable at 3 shots exhibit systematic \textit{illegitimation creep}. This is most pronounced in Olmo, where $\Delta_{\mathrm{LB}}$ surges from an insignificant $+1.58$ ($p = 0.187$) to a massive $+10.38$ ($p < .0001$) when moving from 3 to 5 shots. A similar trend is observed in Gemma, where the directional bias doubles from $+4.00$ to $+8.10$ percentage points. Tables \ref{tab:legitimization_bias_cmr_3shot} and \ref{tab:legitimization_bias_cmr_5shot} display these experimental results for the Cameroon dataset.

AfroConfliLLAMA is the only model in the cohort to trend toward False Legitimation ($V \rightarrow B$) rather than False Illegitimation. In the 3-shot setting, it recorded a $\Delta_{\mathrm{LB}}$ of $-1.05$ percentage points, which shifted toward even greater neutrality at 5 shots ($\Delta_{\mathrm{LB}} = -0.48$). Crucially, in both instances, the $p$-values ($0.238$ and $0.605$, respectively) fail to reject the null hypothesis of directional neutrality. This indicates that for a fine-tuned model, In-Context Learning acts as a refinement tool rather than a source of new bias.

Taken together, these results indicate that ICL does not constitute a general solution for bias elimination. Rather than removing bias, few-shot prompting appears to shift its source from model-internal representations to prompt-induced ones. ICL should therefore be understood as a mechanism for directional steering that, in the absence of fine-tuning, risks amplifying the very normative biases it is intended to correct.

\subsection{Error Trace Analysis}

To better understand the drivers of the models' vulnerability to semantic shifts, we analyzed the explainability traces of the top 20 low-ambiguity events on which model disagreed. For open-weight models and AfroConfliLLAMA, we utilized Rationale-Flip Concordance (RFC) to measure the alignment between a label flip and the model's stated reasoning, obtained via Chain-of-Thought (CoT) prompting ~\cite{10.5555/3600270.3602070}. For AfroConfliBERT, we applied Layer Integrated Gradients (LIG) to isolate the primary token attributions driving the classification. See section \ref{appendix:rationale_extraction_integrated_gradients_attribution} of the appendix for more details.

The RFC metrics shown in Table \ref{tab:error_trace_detailed_cmr} demonstrate a profound disconnect between the open-weight models' classification behavior and their stated reasoning. When models like Llama and Gemma encounter a semantic perturbation, they reliably alter their stated rationale in 97.7\% and 94.4\% of cases, respectively. However, this reasoning is rarely faithful to the input text: their concordant score, which measures the rate at which the model explicitly cites the injected perturbation as the cause of the flip, is only 18.2\% and 16.7\% respectively. The divergence is most extreme in Mistral, which generated a new rationale in 81.2\% of its label flips, but successfully anchored that reasoning to the injected word only 8.3\% of the time. This indicates that open weight models are not analytically responding to the text. Following the framework of Turpin et al. ~\cite{turpin2023language}, this high rate of unfaithful rationalization on simple, structurally clear tasks is a definitive marker of model-induced bias. We hypothesize that subjective framing may act as a latent trigger, inducing post-hoc confabulation that masks systemic normative biases beneath the guise of logical reasoning.
The LIG analysis of the domain-specialized AfroConfliBERT model clarifies the mechanical basis of its high perturbation resilience. In 60\% of the evaluated events, the single highest-attributed token driving the model’s prediction was a core kinetic verb (e.g., \textit{killed}, \textit{kidnapped}, \textit{ambushed}, \textit{attack}). This confirms a strong semantic anchoring to objective factual actions, validating its immunity to subjective adjectives. However, the data also reveals a latent structural vulnerability: in 40\% of cases, the model’s primary attribution anchor was mere punctuation (specifically '\textit{.}' and '\textit{)}'). While the fine-tuned architecture successfully filters out normative framing, its outsized reliance on formatting exposes a partial \textbf{Clever Hans} effect \cite{Pacchiardi2024}, suggesting that its stability may be derived from syntactic memorization of the training distribution rather than pure semantic comprehension.

\begin{table}[t]
\centering
\caption{Legitimization error rates for 3-shot LLM prompts on the Cameroon test set
($|V| = 362$ true violence against civilians, $|B| = 339$ true battles).
$n_{\mathrm{FL}}$ denotes V$\rightarrow$B errors (false legitimation) and
$n_{\mathrm{FI}}$ denotes B$\rightarrow$V errors (false illegitimation).
$\varepsilon$ denotes error rates in percent; 95\% confidence intervals are Wilson score intervals.
$\Delta_{\mathrm{LB}} = \varepsilon_{\mathrm{FI}} - \varepsilon_{\mathrm{FL}}$ (percentage points).
$p$ reports a two-tailed two-proportion $z$-test.
}
\label{tab:legitimization_bias_cmr_3shot}
\small
\setlength{\tabcolsep}{6pt}
\renewcommand{\arraystretch}{1.15}

\begin{tabular}{l r r l r r l r r}
\toprule
\textbf{Model} &
$n_{\mathrm{FL}}$ &
$\varepsilon_{\mathrm{FL}}$ (\%) &
95\% CI$_{\mathrm{FL}}$ &
$n_{\mathrm{FI}}$ &
$\varepsilon_{\mathrm{FI}}$ (\%) &
95\% CI$_{\mathrm{FI}}$ &
$\Delta_{\mathrm{LB}}$ (pp) &
$p$ \\
\midrule
AfroConfliLLAMA & 7 & 1.93 & [0.94, 3.94] & 3 & 0.88 & [0.30, 2.57] & -1.05 & 0.238 \\
Gemma & 9 & 2.49 & [1.31, 4.67] & 22 & 6.49 & [4.32, 9.63] & +4.00 & 0.010 \\
Llama & 15 & 4.14 & [2.53, 6.72] & 19 & 5.60 & [3.61, 8.60] & +1.46 & 0.380 \\
Mistral & 1 & 0.28 & [0.05, 1.54] & 32 & 9.44 & [6.75, 13.04] & +9.16 & < .0001 \\
Olmo & 6 & 1.66 & [0.76, 3.56] & 11 & 3.24 & [1.82, 5.71] & +1.58 & 0.187 \\
\bottomrule
\end{tabular}
\end{table}

\begin{table}[t]
\centering
\caption{Legitimization error rates for 5-shot LLM prompts on the Cameroon test set
($|V| = 362$ true violence against civilians, $|B| = 339$ true battles).
$n_{\mathrm{FL}}$ denotes V$\rightarrow$B errors (false legitimation) and
$n_{\mathrm{FI}}$ denotes B$\rightarrow$V errors (false illegitimation).
$\varepsilon$ denotes error rates in percent; 95\% confidence intervals are Wilson score intervals.
$\Delta_{\mathrm{LB}} = \varepsilon_{\mathrm{FI}} - \varepsilon_{\mathrm{FL}}$ (percentage points).
$p$ reports a two-tailed two-proportion $z$-test.
}
\label{tab:legitimization_bias_cmr_5shot}
\small
\setlength{\tabcolsep}{6pt}
\renewcommand{\arraystretch}{1.15}

\begin{tabular}{l r r l r r l r r}
\toprule
\textbf{Model} &
$n_{\mathrm{FL}}$ &
$\varepsilon_{\mathrm{FL}}$ (\%) &
95\% CI$_{\mathrm{FL}}$ &
$n_{\mathrm{FI}}$ &
$\varepsilon_{\mathrm{FI}}$ (\%) &
95\% CI$_{\mathrm{FI}}$ &
$\Delta_{\mathrm{LB}}$ (pp) &
$p$ \\
\midrule
AfroConfliLLAMA & 6  & 1.66 & [0.76, 3.56] & 4  & 1.18 & [0.46, 3.00] & -0.48 & 0.605 \\
Llama & 13 & 3.59 & [2.11, 6.04] & 12 & 3.54 & [2.04, 6.07] & -0.05 & 0.970 \\
Gemma & 7  & 1.93 & [0.94, 3.94] & 34 & 10.03 & [7.25, 13.71] & +8.10 & < .0001 \\
Mistral & 1  & 0.28 & [0.05, 1.54] & 36 & 10.62 & [7.75, 14.39] & +10.34 & < .0001 \\
Olmo & 3  & 0.83 & [0.28, 2.40] & 38 & 11.21 & [8.27, 15.02] & +10.38 & < .0001 \\
\bottomrule
\end{tabular}
\end{table}

\begin{table}[t]
\centering
\caption{Error trace summary for the Cameroon test set. 
$N_{\mathrm{Flips}}$ denotes total label flips across all perturbation types. 
$N_{\mathrm{Baseline}}$ denotes flips induced specifically by neutral controls (brittleness). 
$\varepsilon_{\Delta \mathrm{Rat}}$ and $\varepsilon_{\mathrm{RFC}}$ report rationale change and concordance rates respectively. 
For LIG, $N_{\mathrm{Flips}}$ reports total events analyzed for attribution.
}
\label{tab:error_trace_detailed_cmr}
\small
\setlength{\tabcolsep}{6pt}
\renewcommand{\arraystretch}{1.15}

\begin{tabular}{l c r r r r l}
\toprule
\textbf{Model} &
Method &
$N_{\mathrm{Flips}}$ &
$N_{\mathrm{Baseline}}$ &
$\varepsilon_{\Delta \mathrm{Rat}}$ (\%) &
$\varepsilon_{\mathrm{RFC}}$ (\%) &
Primary Anchor \\
\midrule
AfroConfliLLAMA     & RFC & 11 & 2 & 36.4 & 0.0  & --- \\
Gemma               & RFC & 18 & 1 & 94.4 & 16.7 & --- \\
Llama               & RFC & 44 & 4 & 97.7 & 18.2 & --- \\
Mistral             & RFC & 48 & 2 & 81.2 & 8.3  & --- \\
Olmo                & RFC & 68 & 2 & 83.8 & 10.3 & --- \\
\midrule
AfroConfliBERT      & LIG & 20 & ---  & ---  & --- & Verbs (60\%), Punct. (40\%) \\
\bottomrule
\end{tabular}
\end{table}

\section{Conclusion}
\label{sec:conclusion}
This study demonstrates that language models do not provide neutral 
representations of political violence in Nigeria and 
Cameroon. The directional biases we observe 
vary systematically across model families, with vanilla LLMs and domain-adapted architectures each exhibiting distinct and reproducible error profiles.

We identify two distinct failure modes. Open-weight models consistently exhibit False Illegitimation bias, with Gemma misclassifying up to 18.29\% of legitimate battles as civilian-targeted violence while committing zero False Legitimation errors across both countries (Tables~\ref{tab:legitimization_bias_nga} and \ref{tab:legitimization_bias_cmr}). AfroConfliBERT and AfroConfliLLAMA achieve near-directional neutrality, with Legitimization Bias differences statistically indistinguishable from zero. However, domain adaptation does not fully resolve actor-based selection bias: even AfroConfliBERT and AfroConfliLLAMA exhibit significant state-actor neutrality bias, with state actors up to 31.7 percentage points less likely to receive a violence label than non-state actors in identical tactical contexts in Nigeria (Table~\ref{tab:fairness_results_combined}), indicating that fairness-aware fine-tuning objectives are necessary alongside domain adaptation.

Vanilla LLMs are further limited by acute fragility to lexical framing. Delegitimizing phrases drive flip rates of up to 66.7\% in Cameroon and 34.2\% in Nigeria, yet the same perturbations that are statistically significant in one context are not significant in the other (Tables~\ref{tab:word_level_perturbation_sensitivity_nga} and 
\ref{tab:word_level_perturbation_sensitivity_cmr}), suggesting that lexical sensitivity varies by geographic context. AfroConfliBERT and AfroConfliLLAMA are largely impervious to these perturbations, recording near-zero flip rates across all perturbation categories (Tables~\ref{tab:vulnerability_stats_nga} and 
\ref{tab:vulnerability_stats_cmr}). Studies of adversarial robustness in related classification tasks demonstrate similar patterns of vulnerability to semantically-preserving lexical perturbations \citep{yang2024adversarial}, while mechanistic interpretability work on toxicity detection reveals disproportionate reliance on lexical markers rather than contextual meaning \citep{garg2025concept}. This creates an exploitable vulnerability where strategic framing can manipulate automated classifications without altering factual event content.

These failure modes carry asymmetric accountability risks. Systems that over-classify violence preserve visibility of potential violations but risk false alarms. Systems that legitimize state violence through linguistic bias create accountability gaps that consistently favor perpetrators who control information environments. Neither is acceptable for conflict monitoring. No single architecture reconciles directional neutrality, actor-based fairness, and lexical robustness simultaneously. Domain adaptation substantially mitigates directional bias and lexical fragility but does not eliminate actor-based selection bias. Addressing these limitations requires a shift from optimizing raw classification accuracy toward achieving both directional neutrality and actor-based fairness.

We therefore call for three evidence-grounded interventions. First, fairness-aware fine-tuning objectives that explicitly optimize for actor-based parity, building on the domain adaptation gains demonstrated by AfroConfliBERT and AfroConfliLLAMA. Second, mandatory adversarial robustness evaluation against lexical manipulation as a pre-deployment requirement, given that flip rates of up to 66.7\% can be engineered by inserting single phrases without altering factual event content. Third, context-specific human-in-the-loop oversight calibrated to regional difficulty: our selective prediction analysis indicates that even the strongest models require human review for approximately 47\% of cases in Cameroon to maintain an accuracy target above 80\% (Section \ref{subsubsection:Deployment}). Future work should examine whether the geographic divergence in lexical sensitivity we observe extends to other conflict contexts in Africa, and whether fairness-aware fine-tuning objectives can reduce residual actor-based selection bias without compromising the directional neutrality and lexical robustness gains demonstrated by AfroConfliBERT and AfroConfliLLAMA.


\section*{Generative AI Usage Statement}

In accordance with FAccT and ACM policies, the authors declare that generative AI was employed only as a proofreading and copy-editing aid. All core ideas, arguments, and text were drafted manually by the authors. Gemini 3 Flash was used to identify grammatical errors and suggest stylistic improvements to the author-written text. The authors have reviewed and verified all suggestions; the final manuscript remains the original work of the authors, who accept full responsibility for its content.

\bibliographystyle{ACM-Reference-Format}
\bibliography{ref}

\newpage
\appendix

\section{Discussion on per-class model metrics}
\label{appendix:per-class_model_metrics}

\subsection{Model Hierarchy and Specialist Dominance}
Across both geographies, AfroConfliBERT and AfroConfliLLAMA consistently outperform other models. Both models demonstrate robustness, maintaining high $F_1$ scores even on complex labels. In contrast, the vanilla LLMs achieve moderate success on labels \textit{V} and \textit{B} and struggle with the other labels, indicating that they may lack the semantic breadth required for these specific classification tasks compared to fine-tuned models.

\subsection{The Difficult Label Problem (R and S)}
There is a universal performance dip across most models for labels \textit{R} and \textit{S}. This indicates that these categories are likely linguistically ambiguous.

\subsection{Cross-Country Robustness vs. Sensitivity}
\begin{itemize}
    \item \textbf{Nigeria}: Models show more balanced scores across labels, likely due to the higher support (sample size) for labels like V ($N=56$) and P ($N=21$).
    \item \textbf{Cameroon}: Performance becomes highly volatile. For example, Olmo has an $F_1$ of 0.724 for \textit{P} in Nigeria, but this crashes to 0.194 in Cameroon. This suggests that models might be highly sensitive to specific context or dialectal nuances when the sample size is small.
\end{itemize}

\begin{figure}[htbp]
    \centering
    \includegraphics[width=\textwidth]{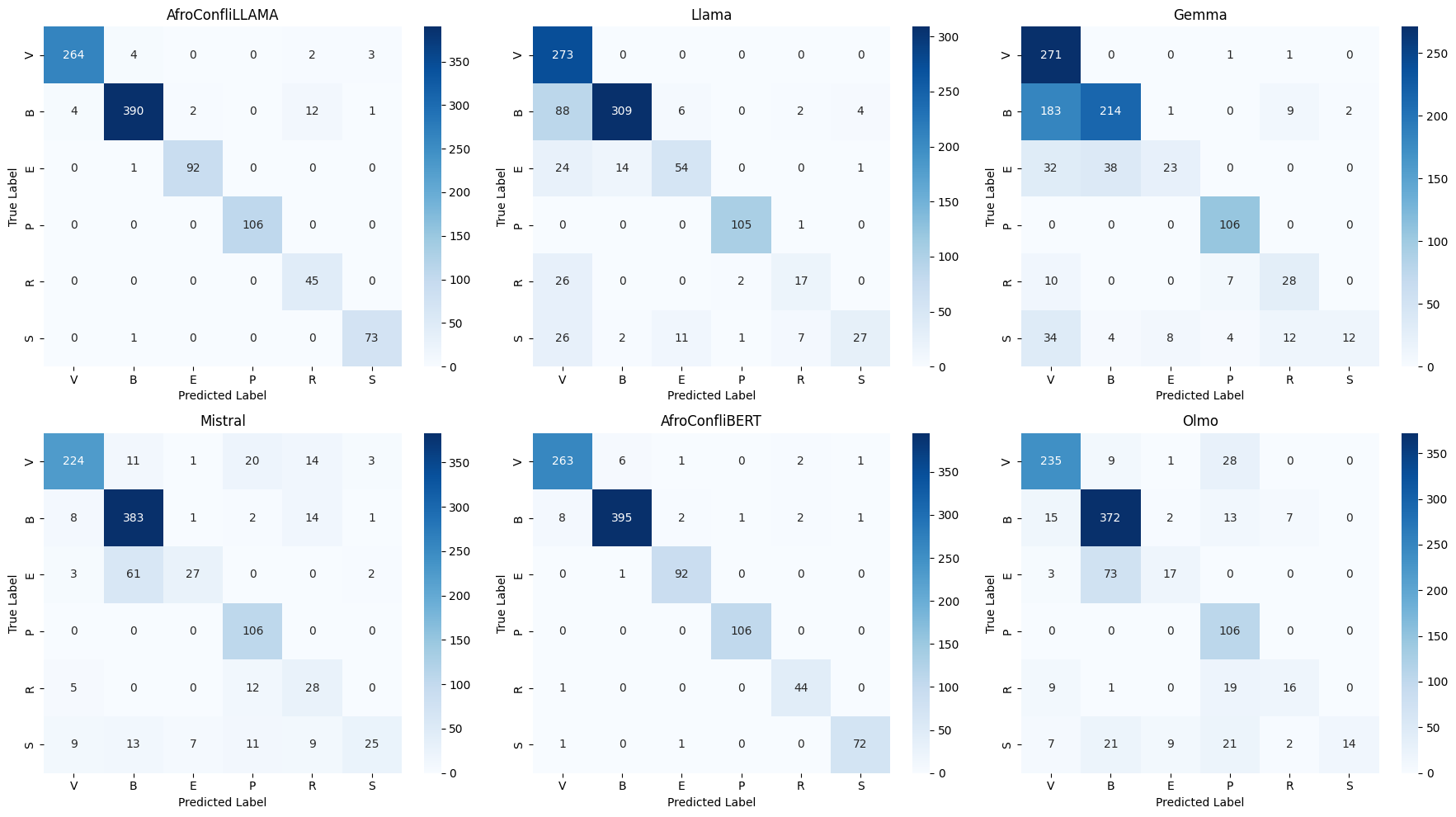}
    \caption{Confusion Matrices for Model Performance on Nigeria}
    \Description[Confusion Matrices for Model Performance on Nigeria]{Confusion Matrices for Model Performance on Nigeria}
    \label{fig:confusion_matrices_nigeria}
\end{figure}

\begin{figure}[htbp]
    \centering
    \includegraphics[width=\textwidth]{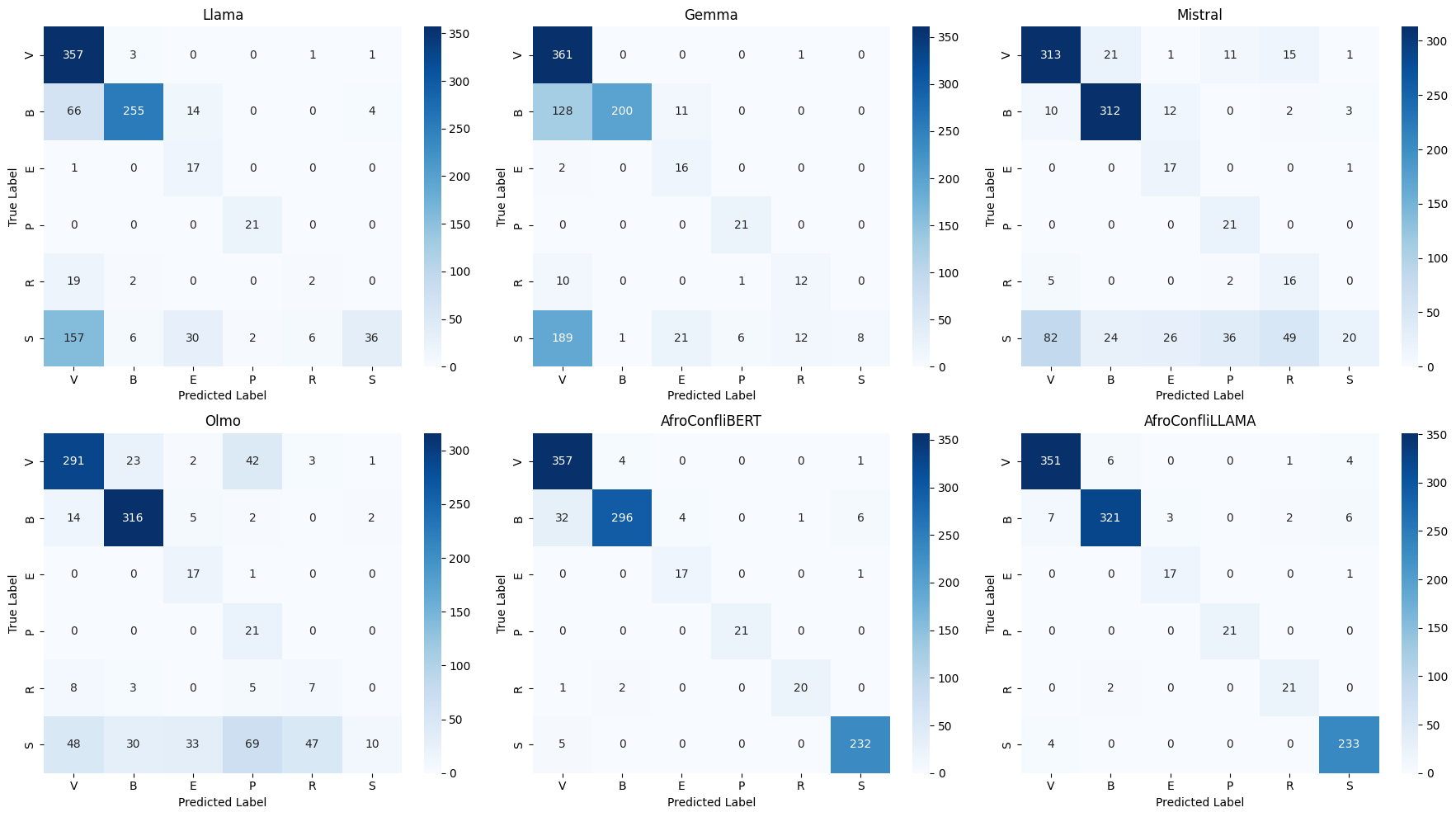}
    \caption{Confusion Matrices for Model Performance on Cameroon}
    \Description[Confusion Matrices for Model Performance on Cameroon]{Confusion Matrices for Model Performance on Cameroon}
    \label{fig:confusion_matrices_cameroon}
\end{figure}

\begin{figure*}[ht]
    \centering
    \includegraphics[width=0.85\textwidth]{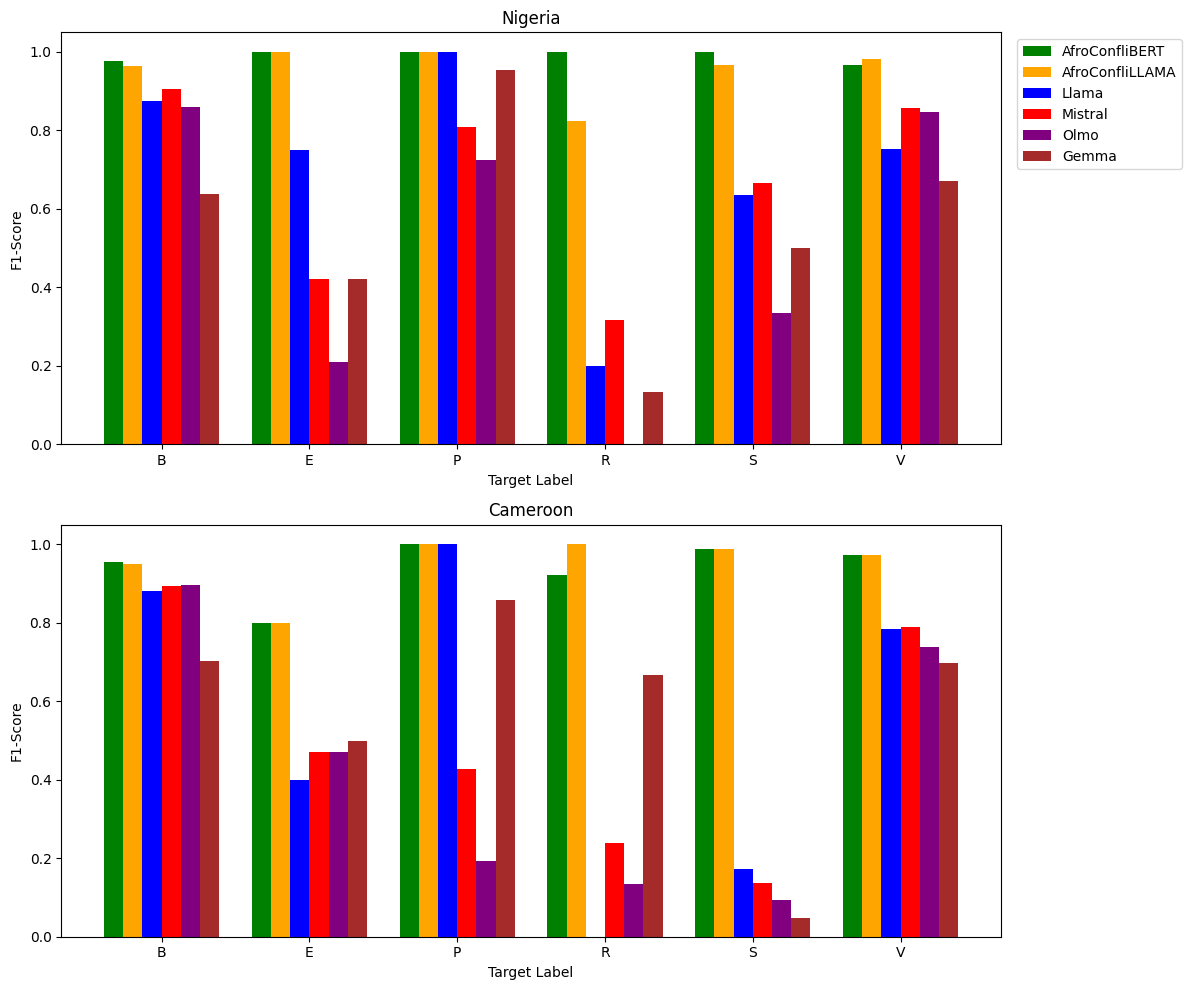}
    \caption{F1-Score comparison across six models for classification labels in Nigeria (top) and Cameroon (bottom).}
    \Description[F1-Score comparison across six models]{F1-Score comparison across six models for classification labels in Nigeria (top) and Cameroon (bottom).}
    \label{fig:f1_comparison_countries_zero_shot}
\end{figure*}

\section{Formal Definitions of Legitimization Metrics}
\label{appendix:metrics}

This section provides the formal mathematical foundations for the metrics used to evaluate model bias. We frame conflict classification not merely as a statistical task, but as a normative assessment aligned with International Humanitarian Law (IHL).

\subsection{Legitimacy Classification Framework}

Let $\mathcal{D} = \{(x_i, y_i)\}_{i=1}^{N}$ denote our evaluation dataset, where $x_i$ represents the $i$-th conflict event description and $y_i \in \{\text{V, B, E, P, R, S}\}$ is the ground-truth label provided by ACLED. Our analysis focuses on the binary distinction between:
\begin{itemize}
    \item \textbf{Violence Against Civilians (V)}: Illegitimate violence targeting non-combatants, prohibited under IHL.
    \item \textbf{Battles (B)}: Legitimate kinetic engagements between organized armed groups.
\end{itemize}

For a model $\mathcal{M}$ with a prediction function $f_{\mathcal{M}} : \mathcal{X} \rightarrow \mathcal{Y}$, we define two principal error modes that carry asymmetric normative consequences:

\begin{definition}[False Legitimation]
A \textit{False Legitimation} (FL) error occurs when an event of violence against civilians is misclassified as a battle:
\begin{equation}
    \text{FL}_i = \mathbbm{1}[y_i = \text{V} \land f_{\mathcal{M}}(x_i) = \text{B}]
\end{equation}
where $\mathbbm{1}[\cdot]$ is the indicator function. FL represents the \textbf{excusal} or \textbf{legitimization} of prohibited violence by re-categorizing it as a legitimate combatant engagement.
\end{definition}

\begin{definition}[False Illegitimation]
A \textit{False Illegitimation} (FI) error occurs when a battle between combatants is misclassified as violence against civilians:
\begin{equation}
    \text{FI}_i = \mathbbm{1}[y_i = \text{B} \land f_{\mathcal{M}}(x_i) = \text{V}]
\end{equation}
FI represents the \textbf{unwarranted condemnation} or \textbf{delegitimization} of recognized military/police conduct of hostilities.
\end{definition}

\subsection{Aggregate Error Rate Computation}

To enable cross-model comparison, we compute normalized error rates relative to the base frequency of each event type in the ground truth:

\begin{equation}
    \varepsilon_{\text{FL}} = \frac{\sum_{i=1}^{N} \text{FL}_i}{\sum_{i=1}^{N} \mathbbm{1}[y_i = \text{V}]} = \frac{n_{\text{FL}}}{|V|}
\end{equation}

\begin{equation}
    \varepsilon_{\text{FI}} = \frac{\sum_{i=1}^{N} \text{FI}_i}{\sum_{i=1}^{N} \mathbbm{1}[y_i = \text{B}]} = \frac{n_{\text{FI}}}{|B|}
\end{equation}

The FL rate ($\varepsilon_{\text{FL}}$) quantifies the model's "blind spot" for civilian victimization, while the FI rate ($\varepsilon_{\text{FI}}$) quantifies the model's tendency to over-report civilian harm in the context of combat.

\subsection{The Legitimization Bias Difference ($\Delta_{\text{LB}}$)}

To quantify the net directional bias of a model's posterior regarding state or non-state actors, we define the \textit{Legitimization Bias Difference}:

\begin{equation}
    \Delta_{\text{LB}} = \varepsilon_{\text{FI}} - \varepsilon_{\text{FL}}
\end{equation}

The value of $\Delta_{\text{LB}}$ identifies the model's normative tendency:
\begin{itemize}
    \item $\Delta_{\text{LB}} > 0$: \textbf{Delegitimization Bias.} The model is more likely to condemn legitimate battles than to excuse civilian targeting. This suggests a "suspicious" posterior against the actor.
    \item $\Delta_{\text{LB}} < 0$: \textbf{Legitimization Bias.} The model is more likely to excuse attacks on civilians than to condemn battles. This suggests a "justificatory" posterior in favor of the actor.
    \item $\Delta_{\text{LB}} \approx 0$: \textbf{Neutral Error Distribution.} The model's errors are symmetric, indicating no directional bias in its legitimacy assessment.
\end{itemize}

Unlike ratio-based metrics (e.g., $\varepsilon_{\text{FI}}/\varepsilon_{\text{FL}}$), $\Delta_{\text{LB}}$ is robust to zero-count errors and measures the net "bias pressure" in percentage points, which is directly interpretable for policy and auditing purposes.

\section{Calibration and Probabilistic Reliability}
\label{appendix:calibration_metrics}

A critical requirement for the deployment of LLMs is not only categorical accuracy but also \textit{probabilistic reliability}. We evaluate the extent to which each model's predicted confidence aligns with its actual empirical frequency using the Brier Score ($BS$), defined as:

\begin{equation}
BS = \frac{1}{N} \sum_{i=1}^{N} (f_i - o_i)^2
\end{equation}

where $f_i$ represents the predicted probability of the positive class and $o_i$ is the actual binary outcome ($0$ or $1$).

\subsection{The Reliability Gap in Raw LLM Outputs}
Our analysis indicates that the vanilla LLMs exhibited a significant reliability gap in their raw outputs. Across both the Nigeria and Cameroon datasets, raw Brier scores were notably high, ranging from $0.128$ to $0.374$. This suggests that the models' internal confidence estimates are initially poorly calibrated for this specific regional taxonomy, often demonstrating overconfidence in incorrect predictions.

\subsection{Comparison of Calibration Techniques}
To address this miscalibration, we compared the effectiveness of two post-hoc calibration methods:
\begin{itemize}
    \item \textbf{Temperature Scaling:} This parametric approach, which scales the logits by a single scalar value, provided negligible improvements (average gain $< 0.001$). This indicates that the miscalibration is not a uniform global bias but rather a more complex, non-linear distortion.
    \item \textbf{Isotonic Calibration:} As a non-parametric approach, Isotonic mapping yielded robust gains, reducing Brier scores by an average of \textbf{27.7\%} in Cameroon and \textbf{17.2\%} in Nigeria. The superior performance of a non-linear method suggest that model biases are class-specific and sensitive to the semantic complexity of regional labels.
\end{itemize}

\subsection{Isotonic Mapping Analysis and Deductions}

The application of isotonic regression to the predicted probabilities of the vanilla LLMs reveal critical insights into their calibration across the Cameroon and Nigeria datasets. By mapping the original predicted probability ($x$) to a calibrated probability ($y$), we observe several distinct behaviors:

\begin{itemize}
    \item \textbf{General Overconfidence}: In both regions, the majority of the mapping curves lie significantly below the $y=x$ (perfect calibration) diagonal. This confirms a systematic overconfidence across models like Mistral and Llama, where the model's internal confidence score is consistently higher than the actual empirical frequency of correct outcomes.
    
    \item \textbf{Calibrated Probability Plateaus}: The isotonic curves exhibit characteristic step-function behavior. Specifically, for Cameroon, Llama shows a long plateau at $y=0.15$ for all $x < 0.8$. This indicates that any confidence score below 80\% is empirically equivalent to a near-zero probability of correctness, suggesting a high rate of false confidence in the lower probability spectrum.

    \item \textbf{Threshold Sensitivity}: The mappings for Nigeria show a \textit{late-onset} calibration curve for Mistral, which remains at $y=0$ until the original confidence exceeds 0.75. This suggests that for certain geopolitical contexts, these models only begin to provide reliable discriminative power at extremely high confidence thresholds.
    
    \item \textbf{Model Specific Divergence:} While Gemma and Llama follow similar monotonic trends, Mistral often requires a much higher original probability to reach a non-zero calibrated value. This implies that the meaning of a 0.8 confidence score is not cross-model compatible and requires individual regional adjustment.
\end{itemize}

\noindent
The deductions suggest that without these isotonic adjustments, the models would provide misleadingly high certainty in their outputs, particularly in the $[0.4, 0.8]$ range, which is here identified as the primary \textit{Overconfident Region}. These mappings are therefore essential for any downstream decision-making tasks to ensure that the probability scores are grounded in regional empirical reality.

\subsection{Geographic and Model-Specific Variations}
A notable geographic divergence was observed in the calibration results. In the Nigeria dataset, Llama achieved the highest reliability ($BS = 0.116$ post-calibration), whereas in Cameroon, Olmo emerged as the leader ($BS = 0.182$). \\
The fact that models required more significant calibration in Cameroon (averaging a 27\% reduction) compared to Nigeria (averaging 17\%) suggests that the models exhibit higher \textit{probabilistic volatility} when dealing with Cameroonian linguistic nuances or lower-support data. As shown in the Brier score analysis, labels with lower sample sizes tend to exacerbate the discrepancy between model confidence and actual performance.

\subsection{Practical Implications}
These results demonstrate that post-hoc calibration is a \textbf{mandatory post-processing step} for using vanilla LLMs in conflict and event classification. Without calibration, the predicted probabilities are misleading, potentially leading to high false-positive rates in automated systems.

\begin{figure*}[t]
    \centering
    \includegraphics[width=0.95\linewidth]{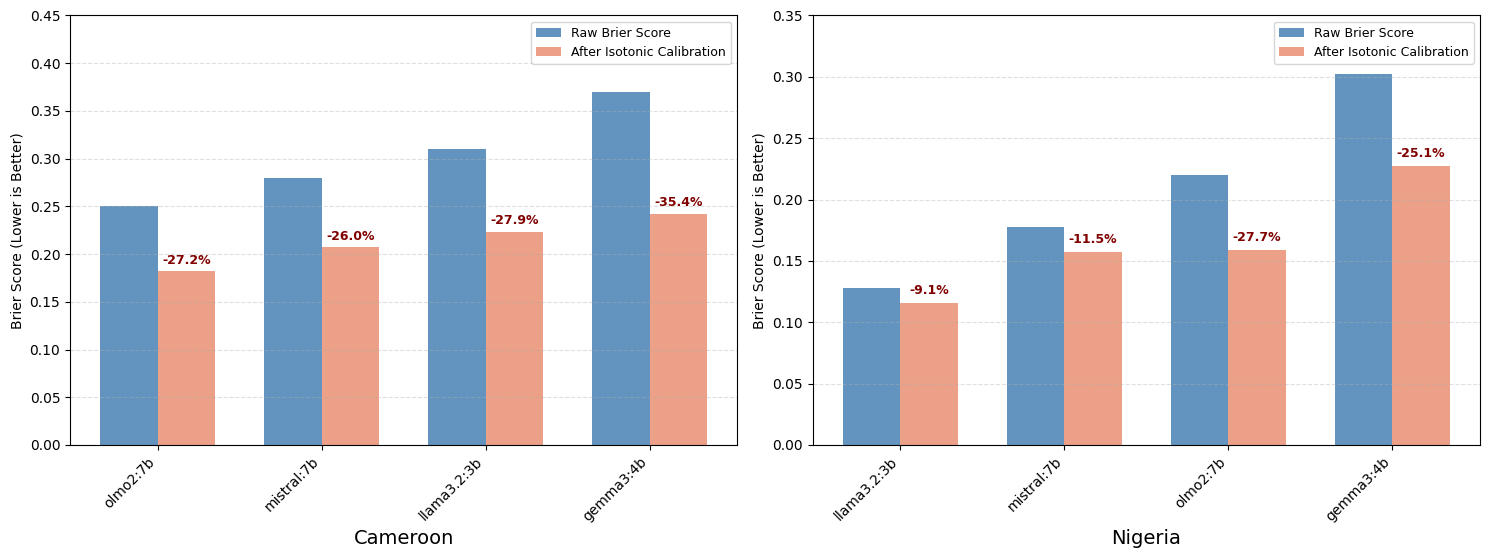}
    \caption{Comparison of Brier Scores before and after Isotonic Calibration for vanilla LLMs. Lower scores indicate better probabilistic calibration. Note the significant reliability gain (up to 35.4\%) across all models after calibration.}
    \Description[Comparison of Brier Scores before and after Isotonic Calibration for vanilla LLMs]{Comparison of Brier Scores before and after Isotonic Calibration for vanilla LLMs. Lower scores indicate better probabilistic calibration. Note the significant reliability gain (up to 35.4\%) across all models after calibration.}
    \label{fig:brier_comparison}
\end{figure*}

\begin{figure*}[t]
    \centering
    \includegraphics[width=0.95\linewidth]{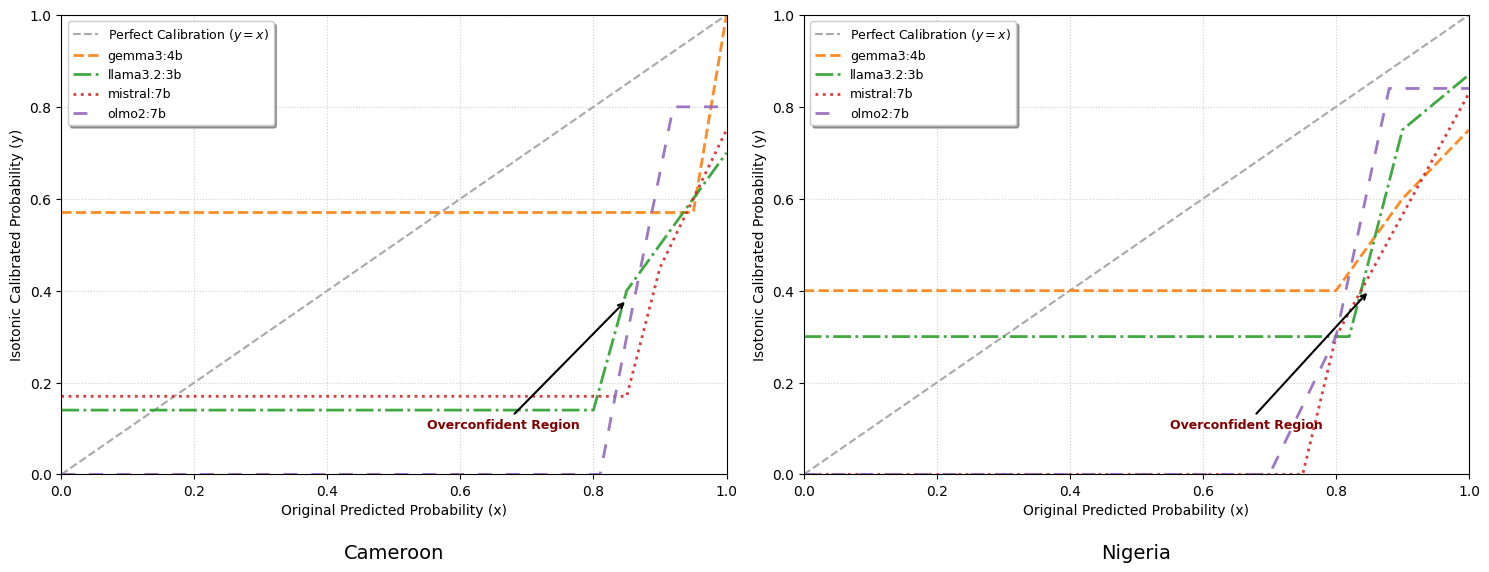}
    \caption{Isotonic mapping functions for vanilla LLMs. The diagonal dashed line ($y=x$) represents a perfectly calibrated model. The curves illustrate the empirical correction applied: most models exhibit significant systematic overconfidence (residing in the lower-right quadrant), requiring a downward adjustment of predicted probabilities to match observed frequencies.}
    \Description[Isotonic mapping functions for vanilla LLMs]{Isotonic mapping functions for vanilla LLMs. The diagonal dashed line ($y=x$) represents a perfectly calibrated model. The curves illustrate the empirical correction applied: most models exhibit significant systematic overconfidence (residing in the lower-right quadrant), requiring a downward adjustment of predicted probabilities to match observed frequencies.}
    \label{fig:calibration_mappings}
\end{figure*}

\subsection{Selective Prediction Analysis}
To evaluate the reliability of model outputs in high-stakes regional contexts, we employ a selective prediction framework. We evaluate the Coverage-Accuracy Tradeoff across three calibration regimes: raw softmax probabilities ($P_{raw}$), Isotonic Regression ($P_{iso}$), and Temperature Scaling ($P_{temp}$). This allows us to determine optimal operating thresholds where the model maintains high precision by abstaining from low-confidence predictions.

\subsubsection{Coverage-Accuracy Tradeoff}
Our analysis reveals a significant divergence in how models handle uncertainty across the Cameroon and Nigeria datasets. We define coverage as the fraction of samples where the model's confidence exceeds a threshold $\tau$, and accuracy as the performance on this filtered subset.

\begin{itemize}
\item \textbf{Model Performance}: In both regions, Llama and Mistral exhibit the most robust selective prediction profiles. For the Nigeria dataset, Llama achieves an accuracy of $\sim$89\% while maintaining 84.5\% coverage ($\tau=0.9$), suggesting that its internal confidence is highly correlated with correctness.
\item \textbf{Calibration}: Raw confidence scores often overestimate accuracy. However, after Isotonic Calibration, the thresholds become more "honest". For example, in the Cameroon context, Mistral maintains a steady accuracy increase as the threshold rises.
\item \textbf{Regional Difficulty}: Accuracy is consistently lower in the Cameroon dataset across all thresholds compared to Nigeria. This suggests that the semantic overlap identified in the confusion matrix analysis (e.g., label \textit{S} vs. \textit{V}) creates an epistemic uncertainty that calibration can identify but not entirely resolve.
\end{itemize}

\noindent
Table \ref{tab:threshold_results} summarizes the performance of the models at two critical operating points: the \textbf{baseline} (zero-abstention) and the \textbf{high-confidence} (precision-optimized) regimes.

\begin{table}[ht]
\centering
\small 
\caption{Vanilla LLM Model Performance at Key Operating Thresholds ($\tau$). Baseline ($\tau=0.0$) represents full coverage, while high-confidence ($\tau=0.9$) demonstrates model reliability under selective prediction. Raw denotes raw softmax probabilities, ISO denotes probabilities calibrated via Isotonic Regression, and TEMP denotes temperature scaled probabilities.}
\label{tab:threshold_results}
\begin{tabular}{@{}llcccccccc@{}}
\toprule
 & & \multicolumn{4}{c}{\textbf{Cameroon}} & \multicolumn{4}{c}{\textbf{Nigeria}} \\
\cmidrule(lr){3-6} \cmidrule(lr){7-10}
\textbf{Model} & \textbf{Method} & \multicolumn{2}{c}{$\tau=0.0$} & \multicolumn{2}{c}{$\tau=0.9$} & \multicolumn{2}{c}{$\tau=0.0$} & \multicolumn{2}{c}{$\tau=0.9$} \\
\cmidrule(lr){3-4} \cmidrule(lr){5-6} \cmidrule(lr){7-8} \cmidrule(lr){9-10}
 & & Cov. & Acc. & Cov. & Acc. & Cov. & Acc. & Cov. & Acc. \\
\midrule
Gemma 3 4B   & ISO  & 1.00 & 0.585 & 0.28 & 0.803 & 1.00 & 0.745 & 0.39 & 0.935 \\
Llama 3.2 3B & TEMP & 1.00 & 0.665 & 0.64 & 0.690 & 1.00 & 0.865 & 0.84 & 0.888 \\
Mistral 7B   & ISO  & 1.00 & 0.695 & 0.53 & 0.811 & 1.00 & 0.800 & 0.56 & 0.946 \\
Olmo 2 7B    & RAW  & 1.00 & 0.575 & 0.45 & 0.777 & 1.00 & 0.810 & 0.51 & 0.892 \\
\bottomrule
\end{tabular}
\end{table}

\subsubsection{Selective Prediction and Model Trustworthiness}
\begin{figure*}[t]
    \centering
    \includegraphics[width=0.95\linewidth]{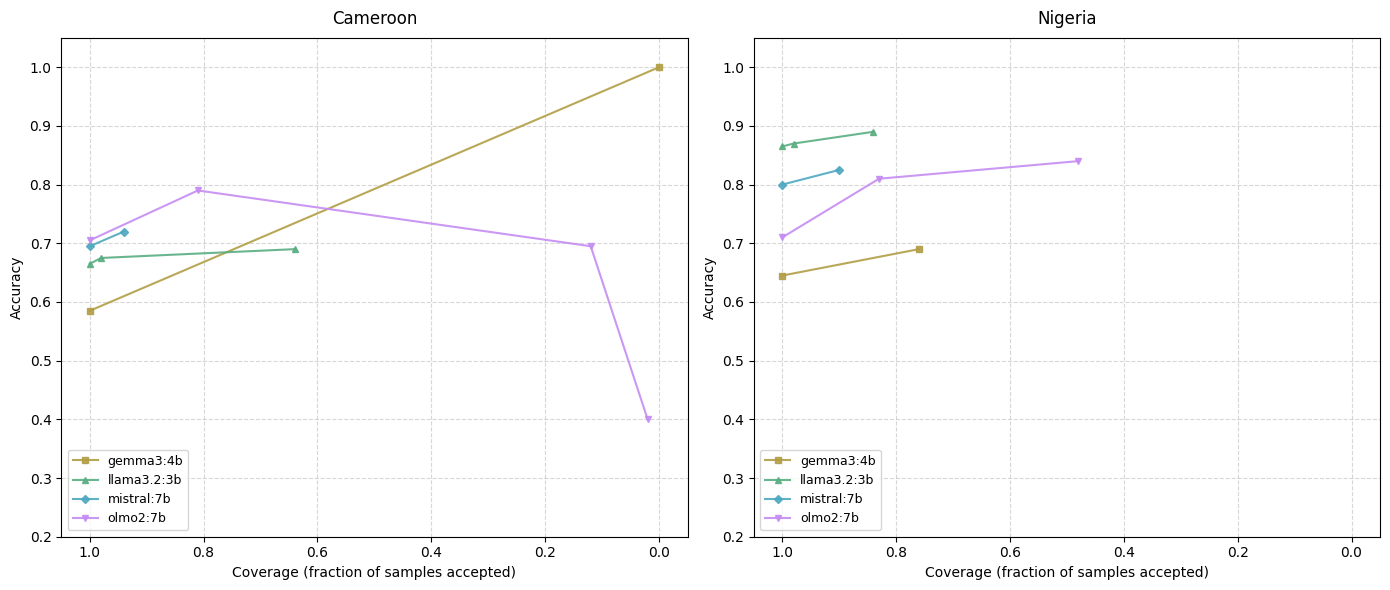}
    \caption{Selective Prediction Profiles: Accuracy vs. Coverage across Cameroon and Nigeria. The x-axis is inverted following academic convention, where moving from left to right represents increasing confidence thresholds ($\tau$) and decreasing coverage. Curves at the top-left of each plot represent the most efficient models (higher accuracy at higher coverage).}
    \Description[Selective Prediction Profiles: Accuracy vs. Coverage across Cameroon and Nigeria]{Selective Prediction Profiles: Accuracy vs. Coverage across Cameroon and Nigeria. The x-axis is inverted following academic convention, where moving from left to right represents increasing confidence thresholds ($\tau$) and decreasing coverage. Curves at the top-left of each plot represent the most efficient models (higher accuracy at higher coverage).}
    \label{fig:selective_prediction_curves}
\end{figure*}

The selective prediction curves in figure \ref{fig:selective_prediction_curves} demonstrate the efficacy of our isotonic calibration across the Cameroon and Nigeria datasets. This visualization represents the \textit{risk-return} profile of the models: how much accuracy is gained by allowing the model to abstain from low-confidence predictions. Across all models, there is a clear monotonic increase in accuracy as the coverage is reduced. In Nigeria, Mistral reaches a precision peak of $>95\%$ as coverage drops below $0.4$, validating that the calibrated confidence scores effectively identify \textit{safe} versus \textit{uncertain} samples.

\subsubsection{Deductions for Deployment}
\label{subsubsection:Deployment}
For researchers and practitioners, these results suggest a multi-tier deployment strategy. In Nigeria, a vanilla Llama can be deployed with a high threshold ($\tau=0.9$) to automate 84\% of tasks with nearly 90\% reliability. Conversely, in Cameroon, the lower accuracy-at-coverage suggests that vanilla LLMs require a human-in-the-loop for roughly 47\% of cases if an accuracy target of $>80\%$ is required.

\section{Fairness Metrics}
\label{appendix:fairness_metrics}

\subsection{Statistical Parity Difference (SPD)}
Statistical Parity requires that the probability of a positive outcome, in our case, the legitimizing label of \textit{Battle} (B), be equal across groups, regardless of the underlying ground-truth distribution. We define the Statistical Parity Difference as:
\begin{equation}
    \text{SPD} = \mathbb{P}(\hat{y} = \text{B} \mid A = g) - \mathbb{P}(\hat{y} = \text{B} \mid A = g')
\end{equation}
\textbf{Intuition:} SPD measures the legitimacy gap. A high absolute SPD suggests that one group is consistently granted a warrior status (legitimacy) more frequently than another, irrespective of whether their actions actually constitute combat or one-sided violence.

\subsection{Equalized Odds: TPR and FPR Parity}
While SPD ensures equal outcomes, it ignores the ground truth ($y$). For high-stakes monitoring, we require \textit{equalized ddds}, which demands that the model's error modes be distributed equitably across groups. This is decomposed into parity of the True Positive Rate (TPR) and False Positive Rate (FPR).

\paragraph{True Positive Rate Difference ($\Delta_{\text{TPR}}$)}
This metric assesses whether the model is equally capable of correctly identifying legitimate combat for both groups:
\begin{equation}
    \Delta_{\text{TPR}} = \mathbb{P}(\hat{y} = \text{B} \mid y = \text{B}, A = g) - \mathbb{P}(\hat{y} = \text{B} \mid y = \text{B}, A = g')
\end{equation}
\textbf{Intuition:} $\Delta_{\text{TPR}}$ measures \textit{Recognition Equality}. A negative $\Delta_{\text{TPR}}$ for a non-state actor suggests that while their kinetic actions are technically battles, the model is blind to their military status compared to state actors.

\paragraph{False Positive Rate Difference ($\Delta_{\text{FPR}}$)}
This metric, which corresponds to the false legitimation rate ($\varepsilon_{\text{FL}}$) parity, assesses the rate at which violence against civilians is incorrectly classified as a battle:
\begin{equation}
    \Delta_{\text{FPR}} = \mathbb{P}(\hat{y} = \text{B} \mid y = \text{V}, A = g) - \mathbb{P}(\hat{y} = \text{B} \mid y = \text{V}, A = g')
\end{equation}
\textbf{Intuition:} $\Delta_{\text{FPR}}$ measures \textit{sanitization bias}. If $\Delta_{\text{FPR}} > 0$, the model is routinely "laundering" the human rights abuses of group $g$ by rebranding their violence against civilians as legitimate combat. This represents a severe failure of accountability, as the model effectively provides algorithmic cover for war crimes committed by specific actors.

\section{Error Correlation with Text Length}
\label{appendix:error_correlation_with_text_length}

We perform a subpopulation error analysis focused on linguistic density (notes length). This analysis identifies whether model performance is sensitive to the amount of contextual information provided in the event descriptions.

\subsection{Analysis across Linguistic Density Slices}
We evaluate model robustness by partitioning the datasets into categorical length quintiles and calculating the Spearman rank correlation between character count and classification error. Tables \ref{tab:error_analysis_nga} and \ref{tab:error_analysis_cmr} detail the error rate deviations ($\Delta \epsilon$) from the model's global mean.

\begin{table}[ht]
\centering
\small
\caption{Error Rate Disparities by Note Length in Nigeria  for vanilla LLMs. $\Delta \epsilon$ represents the deviation from the model's overall error rate. $n$ denotes subpopulation size.}
\label{tab:error_analysis_nga}
\begin{tabular}{@{}lcccc|c@{}}
\toprule
\textbf{Model} &
\multicolumn{4}{c}{\textbf{Error Rate Deviation ($\Delta \epsilon$) by Slice}} &
\textbf{Spearman} \\
& \textbf{Short} ($n{=}303$) & \textbf{Medium} ($n{=}443$) & \textbf{Long} ($n{=}233$) & \textbf{V.\ Long} ($n{=}17$) & \textbf{Correlation} \\
\midrule
Llama 3.2 3B & $+0.020$ & $+0.009$ & $-0.029$ & $-0.158$ & $-0.055$ \\
Gemma 3 4B   & $-0.005$ & $+0.009$ & $-0.004$ & $-0.058$ & $+0.005$ \\
Mistral 7B   & $-0.018$ & $+0.022$ & $-0.007$ & $-0.095$ & $+0.005$ \\
Olmo 2 7B    & $-0.016$ & $+0.012$ & $+0.007$ & $-0.175$ & $+0.005$ \\
\bottomrule
\end{tabular}
\end{table}

\begin{table}[ht]
\centering
\small
\caption{Error Rate Disparities by Note Length in Cameroon for vanilla LLMs. $\Delta \epsilon$ represents the deviation from the model's overall error rate. $n$ denotes subpopulation size.}
\label{tab:error_analysis_cmr}
\begin{tabular}{@{}lcccc|c@{}}
\toprule
\textbf{Model} &
\multicolumn{4}{c}{\textbf{Error Rate Deviation ($\Delta \epsilon$) by Slice}} &
\textbf{Spearman} \\
& \textbf{Short} ($n{=}388$) & \textbf{Medium} ($n{=}470$) & \textbf{Long} ($n{=}138$) & \textbf{V.\ Long} ($n{=}2$) & \textbf{Correlation} \\
\midrule
Llama 3.2 3B & $+0.017$ & $-0.009$ & $-0.008$ & $-0.349$ & $-0.019$ \\
Gemma 3 4B   & $-0.004$ & $+0.010$ & $-0.011$ & $-0.424$ & $+0.012$ \\
Mistral 7B   & $+0.009$ & $-0.004$ & $-0.001$ & $-0.298$ & $-0.025$ \\
Olmo 2 7B    & $+0.015$ & $-0.006$ & $-0.020$ & $-0.310$ & $-0.049$ \\
\bottomrule
\end{tabular}
\end{table}

\subsubsection{Key Diagnostic Insights}
\begin{itemize}
\item \textbf{Performance Loss from Verbosity}: Across both regions, a consistent negative trend is observed as notes progress from \textit{short} to \textit{long}.

\item \textbf{Correlation Weakness:} The Spearman correlations are universally low ($| \rho | < 0.06$ for LLMs). This indicates that while categorical slices show distinct behavioral shifts, there is no simple linear relationship where "longer is always better." The error is driven more by the \textit{quality} and \textit{type} of information in the notes rather than raw character counts.
\end{itemize}

\section{Qualitative Audit of Model Disagreements}
\label{appendix:qualitative_audit_of_disagreements}
To investigate the semantic boundaries where classifications differ, we analyze the events with the highest inter-model disagreements in zero-shot settings, as shown in Tables \ref{tab:nigeria_disagreements} and \ref{tab:cameroon_disagreements}. These represent events where the lexical cues for violence, battles, or strategic developments are ambiguous or contested.

\section{Event Ambiguity Scoring}
\label{appendix:event_ambiguity_scoring}
Without separating model-induced bias from task-inherent uncertainty, a high counterfactual flip rate on, say, \emph{delegitimation\_add} is uninterpretable: you cannot know whether the flip happened because the framing cue pushed the model over a genuine decision boundary, or because the model was already unstable on a clear-cut event for bias-related reasons. The latter is the scientifically interesting finding; the former is expected and uninteresting.

\textbf{Aleatoric vs Epistemic Uncertainty}: Kendall et al. ~\cite{kendall2017uncertainties} distinguishes aleatoric uncertainty from epistemic uncertainty. For conflict classification, genuinely ambiguous events  e.g. unidentified armed groups attacking a village, which ACLED may code as V despite overlapping with B criteria produce aleatoric uncertainty. Framing-sensitive model behaviour on clear-cut events is epistemic. Conflating them inflates our noise floor and dilutes the bias signal.

\textbf{LLMs as Annotators / Item Response Theory (IRT)}: When multiple models classify the same item, the cross-model agreement pattern is a valid proxy for item difficulty ($\beta$ in Item Response Theory). The empirical formula is high label entropy across models + low mean confidence + high confidence variance = high item difficulty. Critically, this is a behavioral signal, not a text heuristic; it is grounded in what the models actually do, not in surface features of the event description. This is the most scientifically credible component of the ambiguity score ~\cite{Gilardi2023, passonneau-carpenter-2014-benefits}.

\textbf{Conflict-specific boundary conditions}: The V/B boundary is the most contested coding decision in ACLED precisely because it requires resolving: (1) whether the non-state actor constitutes an "organized armed group," and (2) whether the civilian population was deliberately targeted. Both are latent judgments, and ACLED notes (the text we use) frequently under-specify them. Events with unidentified actors by definition cannot satisfy the organizational criterion with certainty, producing genuine task ambiguity. The V/B boundary is central in that it reflects the International Humanitarian Law principle of distinction, which permits attacks only against combatants and prohibits targeting civilians; ACLED’s V and B categories therefore map directly onto the legal boundary between lawful and unlawful uses of force ~\cite{Henckaerts_Doswald-Beck_Alvermann_Dormann_Rolle_2005}.

\textbf{Crowd annotation difficulty and reliability}: The disagreement-as-ambiguity framing is validated across decades of annotation research: workers reliably disagree more on genuinely hard items, and this disagreement encodes real information rather than noise. Our models are "crowd workers" in this framework ~\cite{aroyo2015truth}.

Based on the points above, we employ each factor below to construct the event ambiguity score:
\begin{itemize}
    \item \textit{Label entropy}: IRT item difficulty from cross-model label disagreement constitutes 35\% of the score.
    \item \textit{Confidence uncertainty}: This makes up 25\% of the score. A low value means no model "owns" the event.
    \item \textit{Confidence dispersion}: This is given a 20\% weighting. If models get pulled in different directions then we are dealing with a boundary case.
    \item \textit{Text Ambiguity}: Captures V/B boundary patterns on ACLED and is weighted at 20\%.
\end{itemize}

Behavioural dimensions, which come from the models, get 80\% of the weight, while text heuristics get 20\%. This is the correct ordering: what models do on an event is more probative than surface words in the description ~\cite{aroyo2015truth}. With this scoring metric, events register scores between 0 and 1, and are classified into three tiers: high-ambiguity events have scores of at least 0.6, medium-ambiguity events have scores between 0.3 and 0.6, and low-ambiguity events have scores less than 0.3.

\section{Neutral Controls}
\label{appendix:neutral_controls}
Without a semantically neutral baseline, any non-zero CFR is uninterpretable: you cannot tell whether a flip rate of, say, 10\% reflects genuine model sensitivity to framing, or whether 10\% is simply the noise floor of the model on any lexical change whatsoever. This is a standard problem in NLP sensitivity analysis. The canonical solutions come from Ribeiro et al. ~\cite{ribeiro-etal-2020-beyond} which uses invariance tests, minimal edits that must not change the label as a sensitivity floor, Gardner et al. ~\cite{gardner-etal-2020-evaluating} which proposes label-preserving edits to establish the model's local decision boundary and Goel et al. ~\cite{goel-etal-2021-robustness} which argues that any sensitivity experiment needs a neutral substitution arm to calibrate signal-vs-noise. To this end, we add four swap families for the neutral controls, with at most one substitution per family per event: \emph{reporting verbs} e.g. "said" $\rightarrow$ "stated", \emph{day of week} e.g. "Monday" $\rightarrow$ "Tuesday", \emph{temporal connectives} e.g. "later" $\rightarrow$ "afterwards" and \emph{cardinal numerals} e.g. "two" $\rightarrow$ "three".

\section{Rationale Extraction and Integrated Gradients Attribution}
\label{appendix:rationale_extraction_integrated_gradients_attribution}

\paragraph{Rationale extraction:} During counterfactual analysis, we run our explainable prompting strategy on both the original event text and on each perturbed variant that produced a label flip. Our explainable strategy enforces structured factual claims (actor/action/category\_rationale), which Ye \& Durrett ~\cite{ye2022unreliability} show is the most reliable self-explanation form. The key output is not just the rationale text: it is the rationale-flip concordance: did the stated actor/action change when the label flipped? If yes, the model has some coherent evidential path. If no (label flipped but stated rationale is unchanged), that is a confabulation indicator consistent with Turpin et al. ~\cite{turpin2023language}.

\paragraph{Integrated Gradients attribution:} For each disagreement event processed in counterfactual analysis, we compute token-level attribution scores using Captum's \texttt{IntegratedGradients} against the predicted class and we output a per-token importance vector alongside the flip metrics. For an encoder classifier, we cannot ask "why did you flip?" in words. But IG attribution tells us which input tokens most influenced the logit for the predicted class. For flipped perturbations we can compare the attribution on the original vs. perturbed text and identify which token substitution redistributed attention the most. This is the cleanest error-tracing method available for AfroConfliBERT ~\cite{sundararajan2017axiomatic, ferrando2024primer}.

\section{Experimental Reproducibility}
\label{appendix:experimental_reproducibility}

See Table \ref{tab:generation_params} for the full list of generation parameters.

\subsection{Prompting Strategies}

\begin{lstlisting}[language=Python]
class ZeroShotStrategy(PromptingStrategy):
    """Zero-shot direct classification without examples.
    
    This is the current default approach used in the repository.
    It provides category descriptions and asks for direct classification.
    """
    
    def make_prompt(self, event_note: str) -> str:
        """Generate zero-shot classification prompt.
        
        Args:
            event_note: Event description text to classify
            
        Returns:
            Formatted prompt requesting direct classification
        """
        return f"""You are an expert political conflict event analyst.

Classify the following event into one of six categories: {event_note}

Categories (use ONLY these single-letter codes):
- V = Violence against civilians
- B = Battles
- E = Explosions/Remote violence
- P = Protests
- R = Riots
- S = Strategic developments

Return ONLY valid JSON with this structure:
{{
    "label": "<V, B, E, P, R, or S>",
    "confidence": <decimal between 0 and 1>,
    "logits": {{"V": <num>, "B": <num>, "E": <num>, 
                "P": <num>, "R": <num>, "S": <num>}}
}}

CRITICAL: The "label" field must be exactly one of: V, B, E, P, R, S
Do not use numbers, full words, or any other values.

Additional requirements for `logits`:
- Each value in the `logits` object must be a decimal probability between 0 and 1.
- The six logits (V, B, E, P, R, S) must sum to 1.0.
"""
    
    def get_schema(self) -> Dict[str, Any]:
        """Get JSON schema for zero-shot responses.
        
        Returns:
            JSON schema expecting label, confidence, and optional logits
        """
        return {
            "type": "object",
            "properties": {
                "label": {"type": "string", "enum": ["V", "B", "E", "P", "R", "S"]},
                "confidence": {"type": "number"},
                "logits": {
                    "type": "object",
                    "properties": {
                        "V": {"type": "number"},
                        "B": {"type": "number"},
                        "E": {"type": "number"},
                        "P": {"type": "number"},
                        "R": {"type": "number"},
                        "S": {"type": "number"}
                    }
                }
            },
            "required": ["label", "confidence"]
        }
    
    def get_system_message(self) -> Optional[str]:
        """Get system message for zero-shot.
        
        Returns:
            None (zero-shot doesn't use system messages)
        """
        return None
\end{lstlisting}

\textbf{NB:} Our few-shot examples are drawn from the Cameroon and Nigerian datasets. They have low event-ambiguity scores and are correctly classified so models can easily learn conflict distinctions.

\begin{lstlisting}[language=Python]
class FewShotStrategy(PromptingStrategy):
    """Few-shot classification with example demonstrations.
    
    Provides example(s) per category to guide the model's classification.
    """
    
    def __init__(self, config: Optional[Dict[str, Any]] = None):
        """Initialize few-shot strategy.
        
        Args:
            config: Configuration with optional 'examples_per_category' key
                   specifying number of examples to show (1-5, default: 1)
        """
        super().__init__(config)
        self.examples_per_category = config.get('examples_per_category', 1) if config else 1
        if not 1 <= self.examples_per_category <= 5:
            raise ValueError("examples_per_category must be between 1 and 5")
    
    def _get_examples(self, n_per_category: int) -> str:
        """Generate example classifications for few-shot learning.
        
        Args:
            n_per_category: Number of examples per category (1-5)
            
        Returns:
            Formatted string with example input/output pairs
        """
        example_pool = {
            "V": [
                ("Military forces shot and injured a woman in Nongomadiba when they fired shots at a building they believed to be holding Ambazonian Separatists.", "V", 0.89, {"V": 0.89, "B": 0.05, "E": 0.02, "P": 0.01, "R": 0.02, "S": 0.01}),
                ("Security forces opened fire on civilians during a raid in Bamenda, killing two people.", "V", 0.94, {"V": 0.94, "B": 0.03, "E": 0.01, "P": 0.01, "R": 0.01, "S": 0.00}),
                ("Armed soldiers beat and detained three civilians suspected of supporting separatists in Kumbo.", "V", 0.86, {"V": 0.86, "B": 0.02, "E": 0.01, "P": 0.02, "R": 0.03, "S": 0.06}),
                ("Police forces tortured detainees at a checkpoint in Buea, injuring five individuals.", "V", 0.91, {"V": 0.91, "B": 0.03, "E": 0.01, "P": 0.01, "R": 0.02, "S": 0.02}),
                ("Military personnel looted civilian homes and assaulted residents in Mamfe.", "V", 0.87, {"V": 0.87, "B": 0.02, "E": 0.01, "P": 0.02, "R": 0.05, "S": 0.03}),
            ],
            "B": [
                ("The police forces killed one suspected Boko Haram fighter and arrested another in Aissa Karde village.", "B", 0.82, {"V": 0.08, "B": 0.82, "E": 0.03, "P": 0.01, "R": 0.02, "S": 0.04}),
                ("Military forces clashed with separatist fighters in Belo, resulting in casualties on both sides.", "B", 0.93, {"V": 0.03, "B": 0.93, "E": 0.02, "P": 0.01, "R": 0.01, "S": 0.00}),
                ("Government troops engaged Boko Haram militants near Fotokol, killing several insurgents.", "B", 0.91, {"V": 0.04, "B": 0.91, "E": 0.03, "P": 0.00, "R": 0.01, "S": 0.01}),
                ("Armed forces exchanged fire with rebel groups in the Northwest region for several hours.", "B", 0.89, {"V": 0.05, "B": 0.89, "E": 0.03, "P": 0.01, "R": 0.01, "S": 0.01}),
                ("Security forces raided a separatist hideout in Kumba, killing three fighters.", "B", 0.85, {"V": 0.06, "B": 0.85, "E": 0.02, "P": 0.01, "R": 0.02, "S": 0.04}),
            ],
            "E": [
                ("An IED planted by suspected Ambazonian separatists detonated in Matezen village, Santa subdivision, injuring three people.", "E", 0.96, {"V": 0.02, "B": 0.01, "E": 0.96, "P": 0.00, "R": 0.01, "S": 0.00}),
                ("A roadside bomb exploded near a military convoy in Kolofata, wounding two soldiers.", "E", 0.95, {"V": 0.02, "B": 0.02, "E": 0.95, "P": 0.00, "R": 0.01, "S": 0.00}),
                ("Unidentified militants launched a mortar attack on a police station in Mora.", "E", 0.92, {"V": 0.03, "B": 0.03, "E": 0.92, "P": 0.00, "R": 0.01, "S": 0.01}),
                ("An explosive device detonated at a market in Maroua, killing one civilian and injuring ten.", "E", 0.94, {"V": 0.03, "B": 0.01, "E": 0.94, "P": 0.01, "R": 0.01, "S": 0.00}),
                ("Suspected insurgents fired rockets at a military base in the Far North region.", "E", 0.93, {"V": 0.02, "B": 0.03, "E": 0.93, "P": 0.00, "R": 0.01, "S": 0.01}),
            ],
            "P": [
                ("About a hundred residents demonstrated in Djoum town against the government's delay in compensating them after destroying their houses to build the Bikouna-Djoum road.", "P", 0.88, {"V": 0.02, "B": 0.01, "E": 0.01, "P": 0.88, "R": 0.06, "S": 0.02}),
                ("Teachers held a peaceful protest in Yaounde demanding better salaries and working conditions.", "P", 0.95, {"V": 0.01, "B": 0.00, "E": 0.00, "P": 0.95, "R": 0.03, "S": 0.01}),
                ("Students marched through Douala to protest tuition fee increases at public universities.", "P", 0.93, {"V": 0.01, "B": 0.01, "E": 0.00, "P": 0.93, "R": 0.04, "S": 0.01}),
                ("Civil society groups organized a demonstration in Bamenda calling for dialogue and peace.", "P", 0.91, {"V": 0.02, "B": 0.01, "E": 0.01, "P": 0.91, "R": 0.04, "S": 0.01}),
                ("Healthcare workers staged a sit-in at the Ministry of Health demanding payment of arrears.", "P", 0.90, {"V": 0.01, "B": 0.01, "E": 0.00, "P": 0.90, "R": 0.05, "S": 0.03}),
            ],
            "R": [
                ("Residents beat and killed 1 civilian from Ngouma in Tchika, accusing the victim of witchcraft.", "R", 0.79, {"V": 0.12, "B": 0.03, "E": 0.01, "P": 0.03, "R": 0.79, "S": 0.02}),
                ("A mob attacked and burned shops owned by foreigners in Garoua following a dispute.", "R", 0.84, {"V": 0.08, "B": 0.02, "E": 0.02, "P": 0.02, "R": 0.84, "S": 0.02}),
                ("Angry youths vandalized government buildings in Buea after a controversial election result.", "R", 0.81, {"V": 0.05, "B": 0.02, "E": 0.02, "P": 0.08, "R": 0.81, "S": 0.02}),
                ("Residents clashed with police in Edea, destroying vehicles and blocking roads.", "R", 0.83, {"V": 0.06, "B": 0.04, "E": 0.01, "P": 0.04, "R": 0.83, "S": 0.02}),
                ("A violent crowd looted stores and set fire to a police post in Nkongsamba.", "R", 0.80, {"V": 0.09, "B": 0.03, "E": 0.02, "P": 0.03, "R": 0.80, "S": 0.03}),
            ],
            "S": [
                ("Military forces arrested several civilians suspected of connection with ISWAP or Boko Haram militants in Djakana.", "S", 0.77, {"V": 0.10, "B": 0.05, "E": 0.02, "P": 0.02, "R": 0.04, "S": 0.77}),
                ("Government troops increased patrols and established new checkpoints in the Anglophone regions.", "S", 0.85, {"V": 0.04, "B": 0.03, "E": 0.02, "P": 0.02, "R": 0.04, "S": 0.85}),
                ("Security forces conducted a cordon-and-search operation in Mokolo, detaining suspected militants.", "S", 0.81, {"V": 0.07, "B": 0.05, "E": 0.02, "P": 0.01, "R": 0.04, "S": 0.81}),
                ("The army deployed additional personnel to the Far North region to counter insurgent threats.", "S", 0.83, {"V": 0.05, "B": 0.04, "E": 0.03, "P": 0.01, "R": 0.04, "S": 0.83}),
                ("Authorities imposed a curfew in several towns following reports of separatist activity.", "S", 0.79, {"V": 0.06, "B": 0.04, "E": 0.02, "P": 0.03, "R": 0.06, "S": 0.79}),
            ],
        }
        
        import json
        examples_lines = []
        for category in ["V", "B", "E", "P", "R", "S"]:
            for i in range(min(n_per_category, len(example_pool[category]))):
                desc, label, conf, logits = example_pool[category][i]
                output = {"label": label, "confidence": conf, "logits": logits}
                examples_lines.append(f"Event: {desc}")
                examples_lines.append(f"{json.dumps(output)}")
                examples_lines.append("")
        
        return "\n".join(examples_lines)
    
    def make_prompt(self, event_note: str) -> str:
        """Generate few-shot classification prompt with examples."""
        examples = self._get_examples(self.examples_per_category)

        return f"""Examples:

{examples}
--- Now classify the following event in the same format ---
Final Answer: JSON matching the examples above.

Return ONLY valid JSON with this structure:
{{
    "label": "<V, B, E, P, R, or S>",
    "confidence": <decimal between 0 and 1>,
    "logits": {{
        "V": <num>, "B": <num>, "E": <num>,
        "P": <num>, "R": <num>, "S": <num>
    }}
}}

Additional requirements for `logits`:
- Each value must be between 0 and 1.
- The six logits must sum to 1.0.

Event: {event_note}
"""
    
    def get_schema(self) -> Dict[str, Any]:
        return {
            "type": "object",
            "properties": {
                "label": {"type": "string", "enum": ["V", "B", "E", "P", "R", "S"]},
                "confidence": {"type": "number"},
                "logits": {
                    "type": "object",
                    "properties": {
                        "V": {"type": "number"},
                        "B": {"type": "number"},
                        "E": {"type": "number"},
                        "P": {"type": "number"},
                        "R": {"type": "number"},
                        "S": {"type": "number"}
                    }
                }
            },
            "required": ["label", "confidence", "logits"]
        }
    
    def get_system_message(self) -> Optional[str]:
        return """You are an expert political conflict event analyst.
Classify events into one of six categories:
- V = Violence against civilians
- B = Battles
- E = Explosions/Remote violence
- P = Protests
- R = Riots
- S = Strategic developments

Return JSON with label, confidence (0-1), and logits."""
\end{lstlisting}

\begin{lstlisting}[language=Python]
class ExplainableStrategy(PromptingStrategy):
    """Explainable classification with chain-of-thought reasoning.
    
    Asks the model to explain its reasoning before providing classification,
    enabling analysis of how the model arrives at decisions.
    """
    
    def make_prompt(self, event_note: str) -> str:
        """Generate explainable classification prompt with reasoning request."""
        return f"""You are an expert political conflict event analyst.

Classify the following event into one of six categories: {event_note}

Categories (use ONLY these single-letter codes):
- V = Violence against civilians
- B = Battles
- E = Explosions/Remote violence
- P = Protests
- R = Riots
- S = Strategic developments

Step 1 - Brief structured reasoning (exactly three short items):
- Provide three numbered, one-line observations:
  1. Key actors (who)
  2. Key actions (what)
  3. Category rationale (why)
- Each observation must be at most 20 words.
- No extra commentary.

Step 2 - Final answer (valid JSON only):
Return ONLY valid JSON with this structure:
{{
    "reasoning": [<three strings>],
    "label": "<V, B, E, P, R, or S>",
    "confidence": <decimal between 0 and 1>,
    "logits": {{
        "V": <num>, "B": <num>, "E": <num>,
        "P": <num>, "R": <num>, "S": <num>
    }}
}}

CRITICAL: "label" must be exactly one of V, B, E, P, R, S.

Additional requirements for `logits`:
- Each value must be between 0 and 1.
- The six logits must sum to 1.0.
"""
    
    def get_schema(self) -> Dict[str, Any]:
        return {
            "type": "object",
            "properties": {
                "reasoning": {"type": "array", "items": {"type": "string"}},
                "label": {"type": "string", "enum": ["V", "B", "E", "P", "R", "S"]},
                "confidence": {"type": "number"},
                "logits": {
                    "type": "object",
                    "properties": {
                        "V": {"type": "number"},
                        "B": {"type": "number"},
                        "E": {"type": "number"},
                        "P": {"type": "number"},
                        "R": {"type": "number"},
                        "S": {"type": "number"}
                    }
                }
            },
            "required": ["reasoning", "label", "confidence", "logits"]
        }
    
    def get_system_message(self) -> Optional[str]:
        return ("You are an expert political conflict event analyst. "
                "Always explain your reasoning step-by-step before classification.")
\end{lstlisting}

\begin{table*}[t]
\centering
\small
\begin{tabular}{lll}
\toprule
\textbf{Component} & \textbf{Parameter} & \textbf{Value / Description} \\
\midrule

\multicolumn{3}{l}{\textit{Deterministic Inference}} \\
\quad All Models & Decoding & Greedy (deterministic) \\
\quad Ollama & Temperature & 0.0 \\
\quad HuggingFace & do\_sample & False \\
\quad HuggingFace & temperature, top\_p & 1.0 (ignored when sampling disabled) \\

\midrule
\multicolumn{3}{l}{\textit{Ollama Models}} \\
\quad Models & & Gemma, Olmo, Mistral, Llama, AfroConfliLLAMA \\
\quad API & Endpoint & \texttt{http://localhost:11434/api/chat} \\
\quad Output & Format & JSON schema enforced \\
\quad Timeout & & 60 seconds \\

\midrule
\multicolumn{3}{l}{\textit{HuggingFace Models}} \\
\quad Generation & max\_new\_tokens & 96 (default) \\
\quad Device & & Auto (override via \texttt{HF\_DEVICE}) \\
\quad Precision & & float16 (GPU) / float32 (CPU) \\
\quad Tokenization & & truncation + dynamic padding (PyTorch tensors) \\

\midrule
\multicolumn{3}{l}{\textit{Rationale Generation}} \\
\quad Max Tokens & & 160 (overrides classification length) \\
\quad Format & & JSON: reasoning, label, confidence, logits \\
\quad Structure & & 3 items (who/what/why), $\leq$20 words each \\

\midrule
\multicolumn{3}{l}{\textit{AfroConfliBERT (Classifier)}} \\
\quad Tokenization & & truncation + padding (PyTorch tensors) \\

\midrule
\multicolumn{3}{l}{\textit{Attribution (LIG)}} \\
\quad Steps & n\_steps & 50 \\
\quad Baseline & & all-[PAD] tokens \\
\quad Target Layer & & Embedding layer \\
\quad Aggregation & & Sum over embedding dimension \\
\quad Output & & Top-10 tokens per instance \\

\midrule
\multicolumn{3}{l}{\textit{Few-Shot Configuration}} \\
\quad Examples per Class & & 3 (default, configurable) \\

\midrule
\multicolumn{3}{l}{\textit{Environment Variables}} \\
\quad HF\_MAX\_NEW\_TOKENS & & 96 \\
\quad HF\_MAX\_NEW\_TOKENS\_RATIONALE & & 160 \\
\quad HF\_DEVICE & & auto / cuda / cpu \\
\quad NUM\_EXAMPLES & & 3 \\
\quad STRATEGY & & zero\_shot / few\_shot / explainable \\

\bottomrule
\end{tabular}
\caption{Summary of generation parameters and experimental configuration. All models use deterministic decoding to ensure reproducibility.}
\label{tab:generation_params}
\end{table*}

\begin{table}[ht]
\centering
\begin{threeparttable}
\caption{Counterfactual Perturbation Methodology with Neutral Controls}
\label{tab:perturbation_methodology}
\small
\begin{tabular}{p{3.2cm}p{3.8cm}p{3.8cm}p{3.8cm}}
\toprule
\textbf{Perturbation Type} & \textbf{Original Text} & \textbf{Perturbed Text} & \textbf{Neutral Control} \\
\midrule
\textbf{Negation} 
& An unidentified armed group shot and killed a bike rider 
& An unidentified armed group shot and \textit{did not kill} a bike rider 
& An unidentified armed group shot and killed a bike rider \textit{on Tuesday} \\
\addlinespace

\textbf{Legitimation Framing} 
& Military forces shot at protesters in Bamenda
& Military forces shot at protesters in Bamenda \textit{in self-defense} 
& Military forces \textit{reported} shooting at protesters in Bamenda \\
\addlinespace

\textbf{Delegitimation Framing}
& An unidentified armed group shot and killed a bike rider
& \textit{Unprovoked,} an unidentified armed group shot and killed a bike rider 
& An unidentified armed group shot and killed \textit{one} bike rider \\
\addlinespace

\textbf{Actor Substitution} 
& Boko Haram were repulsed by the \textit{military}
& Boko Haram were repulsed by the \textit{security forces}
& Boko Haram were repulsed by the military \textit{afterwards} \\
\addlinespace

\textbf{Provenance Addition} 
& An unidentified armed group shot and killed a bike rider
& \textit{According to state media,} an unidentified armed group shot and killed a bike rider 
& \textit{It was stated that} an unidentified armed group shot and killed a bike rider \\
\addlinespace

\textbf{Intensity Modification} 
& An unidentified armed group shot and killed a bike rider
& An unidentified armed group shot and \textit{brutally} killed a bike rider 
& An unidentified armed group shot and killed a bike rider \textit{on Wednesday} \\
\addlinespace

\textbf{Decontextualization} 
& Armed group killed a bike rider beside a rice farm in Bamunka (Ngo-Ketunjia)
& Armed group killed a bike rider beside a rice farm in \textit{Location X} 
& Armed group killed a bike rider \textit{subsequently} beside a rice farm in Bamunka (Ngo-Ketunjia) \\
\addlinespace

\textbf{Action Substitution} 
& Military \textit{arrested} opposition members
& Military \textit{detained} opposition members 
& Military arrested \textit{four} opposition members \\
\bottomrule
\end{tabular}
\begin{tablenotes}
\item \textit{Note:} Examples drawn from Cameroon conflict events. Italicized text indicates the modification applied. The \textit{Neutral Control} column introduces semantically invariant noise (e.g., swapping days, numerals, temporal connectives, or reporting verbs) to establish a baseline error rate. Any treatment perturbation whose flip rate substantially exceeds this baseline can be attributed to the specific framing dimension under test rather than general lexical sensitivity.
\end{tablenotes}
\end{threeparttable}
\end{table}

\begin{table}[t]
\centering
\caption{Statistical Details for Word-Level Perturbation Sensitivity on Nigerian events.}
\label{tab:word_level_perturbation_sensitivity_nga}
\scriptsize
\begin{tabular}{@{}l l r r r r r r@{}}
\toprule
\textbf{Word / Phrase} & \textbf{Perturbation Type} & \textbf{N} & \textbf{Flip \%} & \textbf{95\% CI} & \textbf{$\Delta\phi$} & \textbf{$d$} & \textbf{$p$} \\
\midrule
unprovoked & Delegitimation & 114 & 21.1 & (14.0--29.7) & $+16.67$ & 0.53 & < .001 \\
using excessive force & Delegitimation & 114 & 19.3 & (12.5--27.7) & $+14.91$ & 0.49 & < .001 \\
killed & Negation & 54 & 16.7 & (7.9--29.3) & $+12.28$ & 0.42 & 0.014 \\
executed & Action substitution & 54 & 14.8 & (6.6--27.1) & $+10.43$ & 0.37 & 0.028 \\
security forces & Actor substitution & 12 & 16.7 & (2.1--48.4) & $+12.28$ & 0.42 & 0.134 \\
confronted & Action substitution & 12 & 16.7 & (2.1--48.4) & $+12.28$ & 0.42 & 0.134 \\
State-run media reported that & Provenance & 114 & 7.9 & (3.7--14.5) & $+3.51$ & 0.15 & 0.270 \\
military personnel & Actor substitution & 6 & 16.7 & (0.4--64.1) & $+12.28$ & 0.42 & 0.270 \\
authorities & Actor substitution & 6 & 16.7 & (0.4--64.1) & $+12.28$ & 0.42 & 0.270 \\
arrested & Negation & 6 & 16.7 & (0.4--64.1) & $+12.28$ & 0.42 & 0.270 \\
Government officials stated that & Provenance & 6 & 16.7 & (0.4--64.1) & $+12.28$ & 0.42 & 0.270 \\
According to state media, & Provenance & 108 & 7.4 & (3.3--14.1) & $+3.02$ & 0.13 & 0.338 \\
violently & Intensity & 78 & 7.7 & (2.9--16.0) & $+3.31$ & 0.14 & 0.359 \\
security personnel & Actor substitution & 42 & 7.1 & (1.5--19.5) & $+2.76$ & 0.12 & 0.445 \\
armed forces & Actor substitution & 12 & 8.3 & (0.2--38.5) & $+3.95$ & 0.16 & 0.459 \\
government troops & Actor substitution & 12 & 8.3 & (0.2--38.5) & $+3.95$ & 0.16 & 0.459 \\
\midrule
\textit{Neutral Control} & \textit{Baseline Noise} & 114 & 4.4 & (1.4--9.9) & -- & -- & -- \\
\bottomrule
\end{tabular}
\vspace{0.5em}
\\[0.5em]
\small

\vspace{1mm}
\footnotesize
\raggedright
\textit{Note:} Flip \% denotes the rate at which inserting the word or phrase changes the predicted label.
95\% Confidence Intervals are calculated safely using the Clopper-Pearson exact method for binomial proportions.
$\Delta\phi$ denotes shift relative to the baseline in percentage points.
$d$ is Cohen’s effect size.
$p$-values use Pearson's $\chi^2$ (or Fisher's Exact test dynamically if any expected cell frequency $< 5$).
Statistics are aggregated across all models. \textit{N} denotes the number of perturbation instances across all models and events. Substitutions are triggered only when the source term appears in the original text. Entity-specific perturbations (e.g. actor) occur less frequently than structural perturbations (e.g. decontextualization) due to within-sample variations resulting in lower $N$ values and wider confidence intervals.
\end{table}

\begin{table}[t]
\centering
\caption{Legitimization error rates by model on the Nigeria test set
($|V|=273$ true violence events, $|B|=409$ true battle events).
$n_{\mathrm{FL}}$ denotes false legitimization errors (V$\rightarrow$B),
$n_{\mathrm{FI}}$ denotes false incrimination errors (B$\rightarrow$V).
Error rates $\epsilon$ are reported as percentages with 95\% Wilson score
confidence intervals.
$\Delta_{\mathrm{LB}} = \epsilon_{\mathrm{FI}} - \epsilon_{\mathrm{FL}}$
(percentage points).
$p$ denotes the two-tailed two-proportion $z$-test.}
\label{tab:legitimization_bias_nga}
\small
\setlength{\tabcolsep}{6pt}
\renewcommand{\arraystretch}{1.15}

\begin{tabular}{l r r l r r l r r}
\toprule
Model &
$n_{\mathrm{FL}}$ &
$\epsilon_{\mathrm{FL}}$ (\%) &
95\% CI$_{\mathrm{FL}}$ &
$n_{\mathrm{FI}}$ &
$\epsilon_{\mathrm{FI}}$ (\%) &
95\% CI$_{\mathrm{FI}}$ &
$\Delta_{\mathrm{LB}}$ (pp) &
$p$ \\
\midrule

AfroConfliBERT
& 0 & 0.00 & [0.00, 1.36]
& 4 & 0.98 & [0.38, 2.49]
& $+0.98$
& 0.100 \\

Gemma
& 0 & 0.00 & [0.00, 1.36]
& 46 & 11.25 & [8.55, 14.65]
& $+11.25$
& $8.4 \times 10^{-9}$ \\

Llama
& 0 & 0.00 & [0.00, 1.36]
& 25 & 6.11 & [4.18, 8.86]
& $+6.11$
& $3.2 \times 10^{-5}$ \\

Mistral
& 3 & 1.10 & [0.37, 3.17]
& 3 & 0.73 & [0.25, 2.13]
& $-0.37$
& 0.606 \\

Olmo
& 2 & 0.73 & [0.20, 2.63]
& 5 & 1.22 & [0.52, 2.82]
& $+0.49$
& 0.542 \\

AfroConfliLLAMA
& 1 & 0.37 & [0.07, 2.05]
& 1 & 0.24 & [0.04, 1.37]
& $-0.13$
& 0.762 \\
\bottomrule
\end{tabular}

\vspace{1mm}
\footnotesize
\raggedright
\textit{Statistical notes.}
$\epsilon_{\mathrm{FL}} = n_{\mathrm{FL}} / |V|$ and
$\epsilon_{\mathrm{FI}} = n_{\mathrm{FI}} / |B|$.
Confidence intervals are Wilson score intervals ($z=1.96$).
$p$ values are from a two-sided two-proportion $z$-test with pooled variance;
for cells with small counts or zeros, these values should be interpreted as
approximations.
\end{table}

\begin{table}[htbp]
\centering
\caption{Model vulnerability to semantic perturbations for Nigeria. Flip \% denotes the rate at which a perturbation changes the predicted label. $\Delta\phi$ is the mean confidence shift over the baseline. $p$ values reflect comparison of perturbation flip rates against neutral controls (Fisher's exact test). $h$ is Cohen's $h$ characterizing the overall effect size of those proportional differences.}
\label{tab:vulnerability_stats_nga}
\scriptsize
\begin{tabular}{@{}l l r c r r l@{}}
\toprule
\textbf{Perturbation Type} &
\textbf{Model} &
\textbf{Flip \%} &
\textbf{95\% CI} &
\textbf{$\Delta\phi$} &
\textbf{$h$} &
\textbf{$p$} \\
\midrule
\multirow{6}{*}{Legitimation framing}
 & AfroConfliBERT      &   0.0 & (0.0--16.8)   & $-0.10$ &  0.00 & 1.000 \\
 & Gemma      &  15.0 & (3.2--37.9)   & $-0.50$ &  0.33 & 0.605 \\
 & Llama   &   5.0 & (0.1--24.9)   & $+1.53$ & -0.21 & 0.605 \\
 & Mistral     &   5.0 & (0.1--24.9)   & $+0.24$ &  0.45 & 1.000 \\
 & Olmo       &   5.0 & (0.1--24.9)   & $-0.24$ & -0.01 & 1.000 \\
 & AfroConfliLLAMA       &   0.0 & (0.0--16.8)   & $+0.00$ & -0.46 & 0.487 \\
\addlinespace
\multirow{6}{*}{Delegitimation framing}
 & AfroConfliBERT      &   7.9 & (1.7--21.4)   & $-2.10$ &  0.57 & 0.544 \\
 & Gemma      &  18.4 & (7.7--34.3)   & $-0.53$ &  0.42 & 0.247 \\
 & Llama   &  23.7 & (11.4--40.2)  & $-2.11$ &  0.36 & 0.304 \\
 & Mistral     &  26.3 & (13.4--43.1)  & $-0.53$ &  1.08 & 0.022 \\
 & Olmo       &  34.2 & (19.6--51.4)  & $-0.18$ &  0.79 & 0.022 \\
 & AfroConfliLLAMA       &  10.5 & (2.9--24.8)   & $+0.00$ &  0.20 & 0.655 \\
\addlinespace
\multirow{6}{*}{Provenance addition}
 & AfroConfliBERT      &   0.0 & (0.0--6.3)    & $-0.84$ &  0.00 & 1.000 \\
 & Gemma      &   7.0 & (1.9--17.0)   & $+1.23$ &  0.07 & 1.000 \\
 & Llama   &   7.0 & (1.9--17.0)   & $-1.23$ & -0.12 & 0.636 \\
 & Mistral     &   5.3 & (1.1--14.6)   & $-2.81$ &  0.46 & 0.569 \\
 & Olmo       &  14.0 & (6.3--25.8)   & $+0.74$ &  0.31 & 0.436 \\
 & AfroConfliLLAMA       &  10.5 & (4.0--21.5)   & $+0.00$ &  0.20 & 0.672 \\
\addlinespace
\multirow{6}{*}{Intensity modification}
 & AfroConfliBERT      &   2.4 & (0.1--12.6)   & $-3.84$ &  0.31 & 1.000 \\
 & Gemma      &   2.4 & (0.1--12.6)   & $+0.95$ & -0.15 & 0.530 \\
 & Llama   &  11.9 & (4.0--25.6)   & $+1.72$ &  0.04 & 1.000 \\
 & Mistral     &   2.4 & (0.1--12.6)   & $+1.17$ &  0.31 & 1.000 \\
 & Olmo       &  23.8 & (12.1--39.5)  & $+2.57$ &  0.56 & 0.148 \\
 & AfroConfliLLAMA       &   0.0 & (0.0--8.4)    & $+0.00$ & -0.46 & 0.311 \\
\addlinespace
\multirow{6}{*}{Negation}
 & AfroConfliBERT      &   0.0 & (0.0--24.7)   & $-2.75$ &  0.00 & 1.000 \\
 & Gemma      &  15.4 & (1.9--45.4)   & $-1.54$ &  0.34 & 0.552 \\
 & Llama   &  23.1 & (5.0--53.8)   & $-0.24$ &  0.34 & 0.374 \\
 & Mistral     &  23.1 & (5.0--53.8)   & $-0.65$ &  1.00 & 0.058 \\
 & Olmo       &  23.1 & (5.0--53.8)   & $-1.66$ &  0.54 & 0.279 \\
 & AfroConfliLLAMA       &   0.0 & (0.0--24.7)   & $+0.00$ & -0.46 & 1.000 \\
\addlinespace
\multirow{6}{*}{Actor substitution}
 & AfroConfliBERT      &   0.0 & (0.0--11.6)   & $-0.01$ &  0.00 & 1.000 \\
 & Gemma      &   6.7 & (0.8--22.1)   & $-0.33$ &  0.06 & 1.000 \\
 & Llama   &  20.0 & (7.7--38.6)   & $+0.53$ &  0.27 & 0.458 \\
 & Mistral     &   0.0 & (0.0--11.6)   & $+0.40$ &  0.00 & 1.000 \\
 & Olmo       &  16.7 & (5.6--34.7)   & $+1.10$ &  0.38 & 0.384 \\
 & AfroConfliLLAMA       &   0.0 & (0.0--11.6)   & $+0.00$ & -0.46 & 0.388 \\
\addlinespace
\multirow{6}{*}{Action substitution}
 & AfroConfliBERT      &   5.6 & (0.7--18.7)   & $-1.07$ &  0.48 & 0.539 \\
 & Gemma      &   8.3 & (1.8--22.5)   & $+0.00$ &  0.12 & 1.000 \\
 & Llama   &  16.7 & (6.4--32.8)   & $+1.92$ &  0.18 & 0.700 \\
 & Mistral     &   0.0 & (0.0--9.7)    & $-0.26$ &  0.00 & 1.000 \\
 & Olmo       &  11.1 & (3.1--26.1)   & $+1.74$ &  0.22 & 0.649 \\
 & AfroConfliLLAMA       &  11.1 & (3.1--26.1)   & $+0.00$ &  0.22 & 0.649 \\
\addlinespace
\multirow{6}{*}{Decontextualization}
 & AfroConfliBERT      &   0.0 & (0.0--10.3)   & $+0.07$ &  0.00 & 1.000 \\
 & Gemma      &   5.9 & (0.7--19.7)   & $-1.18$ &  0.03 & 1.000 \\
 & Llama   &   0.0 & (0.0--10.3)   & $+0.23$ & -0.66 & 0.124 \\
 & Mistral     &   2.9 & (0.1--15.3)   & $+0.47$ &  0.34 & 1.000 \\
 & Olmo       &   5.9 & (0.7--19.7)   & $+1.12$ &  0.03 & 1.000 \\
 & AfroConfliLLAMA       &   0.0 & (0.0--10.3)   & $+0.00$ & -0.46 & 0.358 \\
\addlinespace
\midrule
\multicolumn{7}{l}{\textbf{Neutral Controls}} \\
 & AfroConfliBERT      &   0.0 & (0.0--17.6) & +0.02 & -- & -- \\
 & Gemma      &   5.3 & (0.1--26.0) & +0.00 & -- & -- \\
 & Llama   &  10.5 & (1.3--33.1) & -0.53 & -- & -- \\
 & Mistral     &   0.0 & (0.0--17.6) & +0.26 & -- & -- \\
 & Olmo       &   5.3 & (0.1--26.0) & -0.26 & -- & -- \\
 & AfroConfliLLAMA       &   5.3 & (0.1--26.0) & +0.00 & -- & -- \\
\bottomrule
\end{tabular}
\end{table}

\begin{table}[t]
\centering
\caption{Top 20 Nigeria conflict events with inter-model disagreements.}
\label{tab:nigeria_disagreements}
\scriptsize
\setlength{\tabcolsep}{3pt}
\renewcommand{\arraystretch}{1.15}

\begin{tabular}{@{}l c p{0.18\linewidth} p{0.32\linewidth}
c c c c c c
c c@{}}
\toprule
Event ID & T & Actor & Event (abridged) &
AfroConfliLLAMA & AfroConfliBERT & Gemma & Llama & Mistral & Olmo &
D & Max \\
\midrule

NIG21935 & B &
Zamfara Communal Militia &
Militia attacked security base in Dan Sadau; 12 security personnel killed (Sep 2021). &
B & B & V & B & B & B &
1 & 1.0 \\

NIG34842 & B &
Military Forces of Nigeria (2023-) &
Soldiers attacked policemen in Sabon Gari; some wounded (Feb 2024). &
B & B & V & B & B & B &
1 & 1.0 \\

NIG16585 & R &
Rioters (Nigeria) &
Youths clashed with police during burial procession in Amakohia; one injured (Mar 2020). &
R & R & V & V & P & P &
4 & 1.0 \\

NIG27104 & V &
Military Forces of Nigeria (2015-2023) &
Soldier fired at crowd during local elections in Gwanki; one girl killed (Sep 2021). &
V & V & V & P & P & V &
2 & 1.0 \\

NIG30329 & B &
Unidentified Cult Militia &
Cult group killed four rivals in Kuchibena; body mutilated (Sep 2021). &
R & V & V & V & V & V &
1 & 1.0 \\

NIG34255 & B &
Police Forces of Nigeria (2023-) &
Police clashed with armed group in Emohua forest; two killed, four rescued (Dec 2023). &
B & B & V & B & B & B &
1 & 1.0 \\

NIG21852 & V &
ISWAP - Lake Chad Faction &
ISWAP attacked Christian town in Borno; church and school burned (Aug 2021). &
V & V & V & V & V & E &
1 & 1.0 \\

NIG18752 & V &
Police Forces of Nigeria (2015-2023) &
Police attacked journalist in Udu LGA over curfew enforcement; no fatalities (Oct 2020). &
V & V & V & P & P & P &
3 & 1.0 \\

NIG38445 & B &
SEC: Supreme Eiye Confraternity &
Buccaneers and Eiye clashed in Abeokuta; one killed (Aug 2024). &
R & B & V & V & R & B &
4 & 1.0 \\

NIG34375 & S &
ISWAP &
ISWAP looted drugs from health center in Mainok, Borno (Dec 2023). &
V & S & V & V & R & E &
3 & 1.0 \\

NIG30514 & B &
Police Forces of Nigeria (2015-2023) &
Police fought Kaduna militia; one militia killed, weapons recovered (Mar 2023). &
B & B & V & B & B & B &
1 & 1.0 \\

NIG34705 & S &
Government of Nigeria (2023-) &
Plateau State imposed 24-hour curfew in Mangu after violence (Jan 2024). &
S & S & V & S & S & S &
1 & 1.0 \\

NIG21210 & B &
Military Forces of Nigeria (2015-2023) &
Task force killed two kidnappers in Koton-Karfe (Jun 2021). &
B & B & V & B & B & B &
1 & 1.0 \\

NIG36383 & S &
Police Forces of Nigeria (2023-) &
Police arrested Kaduna militia kingpin near Abuja-Kaduna road; 48 AK-47s recovered (Jan 2024). &
S & S & V & S & S & B &
2 & 1.0 \\

NIG38569 & B &
Military Forces of Nigeria (2023-) &
Troops ambushed Kaduna militia in Giwa; two killed, weapons recovered (Sep 2024). &
B & B & V & B & B & B &
1 & 1.0 \\

NIG36868 & B &
Military Forces of Nigeria (2023-) &
Troops, hunters, and KSVS clashed with Kogi militia; one killed (Jun 2024). &
B & B & V & B & B & B &
1 & 1.0 \\

NIG22187 & V &
Unidentified Armed Group (Nigeria) &
Armed men shot sporadically enforcing sit-at-home in Owerri; police denied attack (Oct 2021). &
V & V & V & P & P & P &
3 & 1.0 \\

NIG34305 & B &
Unidentified Armed Group (Nigeria) &
Gunmen killed police orderly, abducted judge and driver in Oron; released later (Dec 2023). &
V & V & V & B & V & V &
1 & 1.0 \\

NIG34649 & E &
ISWAP &
ISWAP fired shells at Nigerian Army in Nduva; no fatalities (Jan 2024). &
E & E & B & E & E & B &
2 & 1.0 \\

NIG21924 & B &
Unidentified Armed Group (Nigeria) &
Gunmen killed three police and abducted expatriates in Akukutoru (Sep 2021). &
B & B & V & B & B & B &
1 & 1.0 \\

\bottomrule
\end{tabular}

\vspace{1mm}
\footnotesize
\raggedright
\textit{Notes:} T = true label. Gemma = Gemma3 4B, Llama = Llama3.2 3B, Mistral = Mistral 7B, Olmo = Olmo2 7B. D = number of models disagreeing with the majority label, Max = maximum confidence across models.
\end{table}

\begin{table}[t]
\centering
\caption{Top 20 Cameroon conflict events with inter-model disagreements.}
\label{tab:cameroon_disagreements}
\scriptsize
\setlength{\tabcolsep}{3pt}
\renewcommand{\arraystretch}{1.15}

\begin{tabular}{@{}l c p{0.18\linewidth} p{0.32\linewidth}
c c c c c c
c c@{}}
\toprule
Event ID & T & Actor & Event (abridged) &
AfroConfliLLAMA & AfroConfliBERT & Gemma & Llama & Mistral & Olmo &
D & Max \\
\midrule

CAO9055 & V &
Military Forces of Cameroon &
Military shot and killed a man mistaken for separatist in Besali (Sep 2022). &
V & V & V & V & B & B &
2 & 1.0 \\

CAO3117 & B &
ISWAP/Boko Haram &
Boko Haram repulsed by military; hospital destroyed and 3 motorcycles stolen (Mar 2020). &
B & B & V & B & B & B &
2 & 1.0 \\

CAO8281 & S &
ISWAP/Boko Haram &
Militants looted 2 houses in Waza (Dec 2023). &
S & S & V & V & V & V &
2 & 1.0 \\

CAO10079 & S &
Civilians &
Civilians displaced by fear of retaliation from security forces in Diongo (Feb 2023). &
S & S & V & V & P & V &
3 & 1.0 \\

CAO3188 & V &
Military Forces of Cameroon &
Military killed three former separatists in Balikumbat (Apr 2020). &
B & V & V & V & B & B &
2 & 1.0 \\

CAO13551 & S &
Boko Haram &
11 militants surrendered to military in Madakar (Jun 2024). &
S & S & S & S & S & B &
1 & 1.0 \\

CAO4946 & V &
Ambazonian Separatists &
Separatists kidnapped truck driver and assistant in Bukow (Jan 2021). &
V & V & V & V & V & P &
1 & 1.0 \\

CAO4776 & S &
Military Forces of Cameroon &
Military looted Bengang village; fired in the air (Jul 2020). &
S & S & V & R & R & R &
3 & 1.0 \\

CAO11075 & B &
Military Forces of Cameroon &
Military ambushed separatists in Konye; one injured (Nov 2023). &
B & B & V & B & B & B &
2 & 1.0 \\

CAO10854 & B &
Military Forces of Cameroon &
Military clashed with separatists in Bai Panya; 2 killed, 4 arrested (Dec 2023). &
B & B & V & B & B & B &
2 & 1.0 \\

CAO8295 & S &
Military Forces (Rapid Intervention) &
Military defused IED planted by unidentified armed group in Kumba (Dec 2023). &
S & S & E & E & E & E &
4 & 1.0 \\

CAO4198 & V &
Ambazonian Separatists &
Separatists kidnapped 6–7 students in Mulang, Bamenda (Jan 2021). &
V & V & V & V & V & P &
1 & 1.0 \\

CAO12441 & V &
Military Forces of Cameroon &
State security forces killed two separatists in Kumbo (Jul 2023). &
V & V & V & V & B & V &
1 & 1.0 \\

CAO8757 & S &
Unidentified Armed Group &
Militants looted food, clothing, and bicycle in Ganse (Mar 2024). &
S & S & V & V & V & V &
2 & 1.0 \\

CAO14603 & V &
Ambazonian Separatists &
Separatists beat and injured driver and abducted 4 officials in Bambui; motorbike later recovered (Jun 2024). &
V & V & V & V & B & V &
2 & 1.0 \\

CAO13869 & R &
Rioters &
Youth clashed during wake in Belabo; casualties unknown (Jun 2024). &
R & R & R & B & R & P &
3 & 1.0 \\

CAO7298 & V &
ISWAP/Boko Haram &
Militants abducted 2 herders and stole ox in Zigague (May 2023). &
V & V & V & V & V & B &
1 & 1.0 \\

CAO12794 & S &
Military Forces of Cameroon &
Military burnt down 2 shops in Ndop for failing to open during lockdown (Mar 2022). &
S & S & V & R & R & P &
4 & 1.0 \\

CAO2846 & B &
Military Forces of Cameroon &
Military engaged separatists in Peng; 2 civilians killed (Jan 2020). &
B & B & V & B & B & B &
2 & 1.0 \\

CAO4930 & V &
Ambazonian Separatists &
Separatists kidnapped retired officer in Mankon (Nov 2020). &
V & V & V & V & R & P &
2 & 1.0 \\

\bottomrule
\end{tabular}

\vspace{1mm}
\footnotesize
\raggedright
\textit{Notes:} T = true label. Gemma = Gemma3 4B, Llama = Llama3.2 3B, Mistral = Mistral 7B, Olmo = Olmo2 7B. D = number of models disagreeing with the majority label, Max = maximum confidence across models.
\end{table}

\newpage
\begin{figure*}[t]
    \begin{subfigure}{0.44\textwidth}
        \centering
        \includegraphics[width=\linewidth]{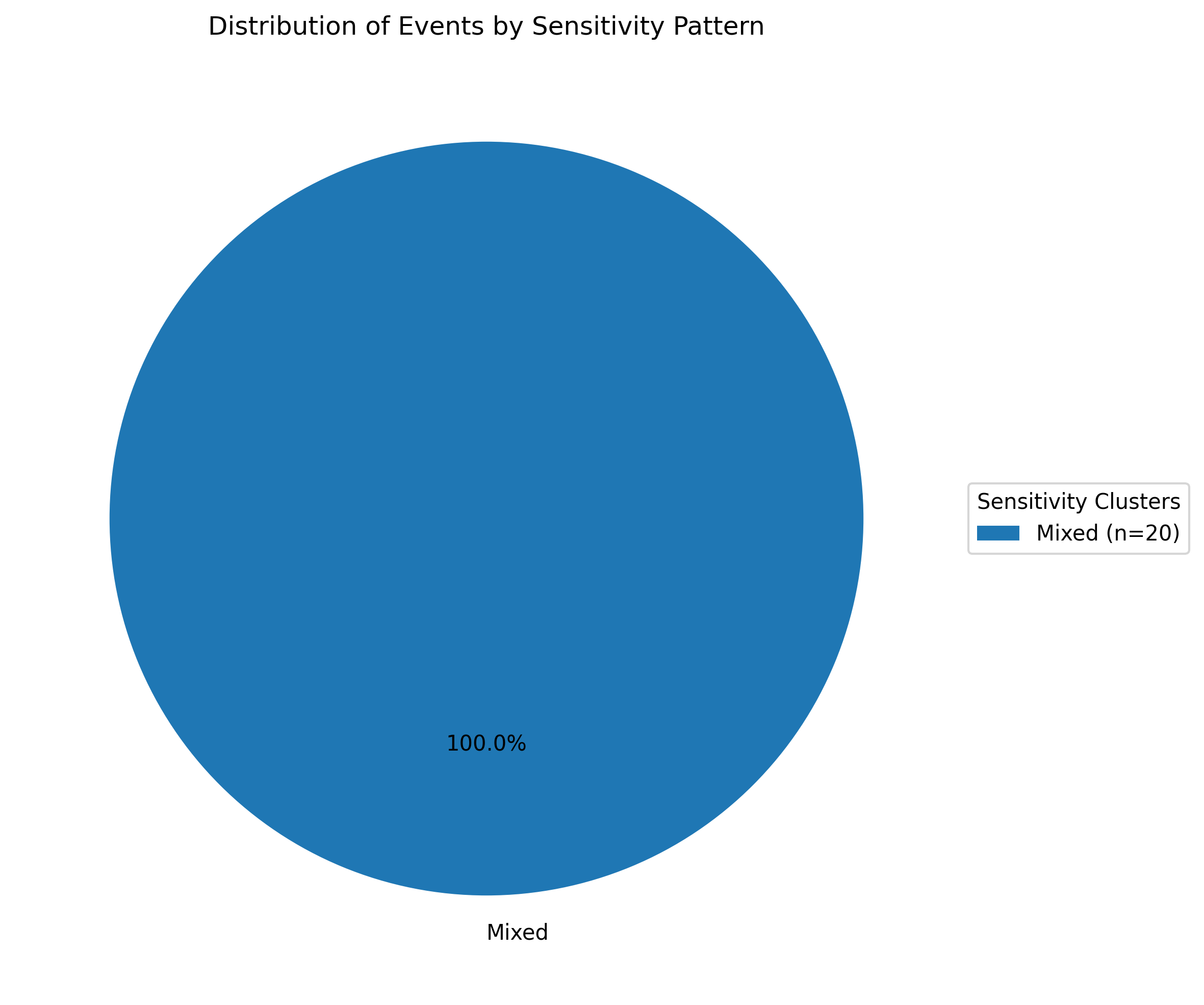}
        \caption{AfroConfliBERT}
        \label{fig:sensitivity_clusters_acledbert}
    \end{subfigure}
    \hfill
    \begin{subfigure}{0.49\textwidth}
        \centering
        \includegraphics[width=\linewidth]{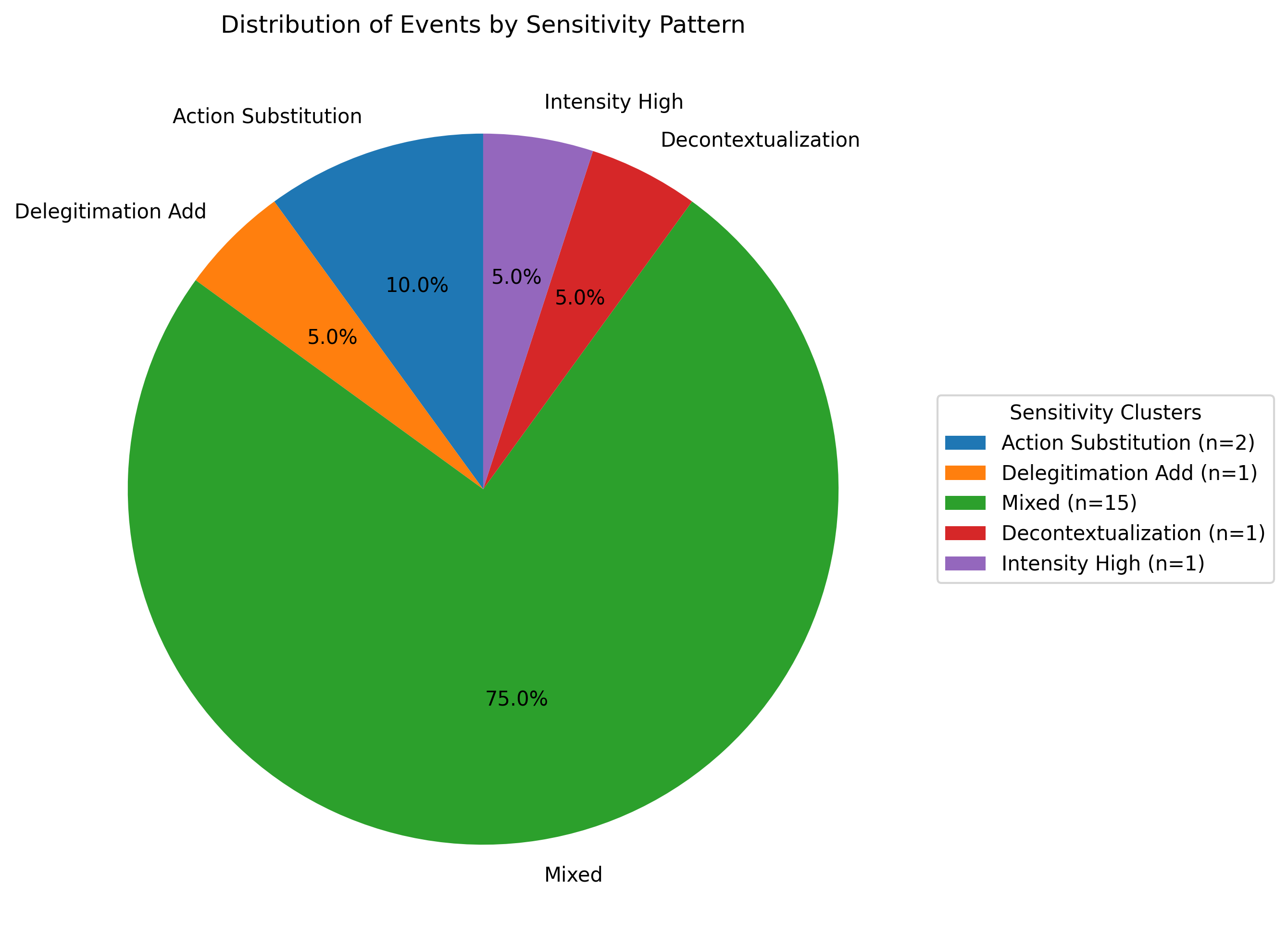}
        \caption{Olmo}
        \label{fig:sensitivity_clusters_olmo}
    \end{subfigure}

    \vspace{0.3cm}

    \begin{subfigure}{0.49\textwidth}
        \centering
        \includegraphics[width=\linewidth]{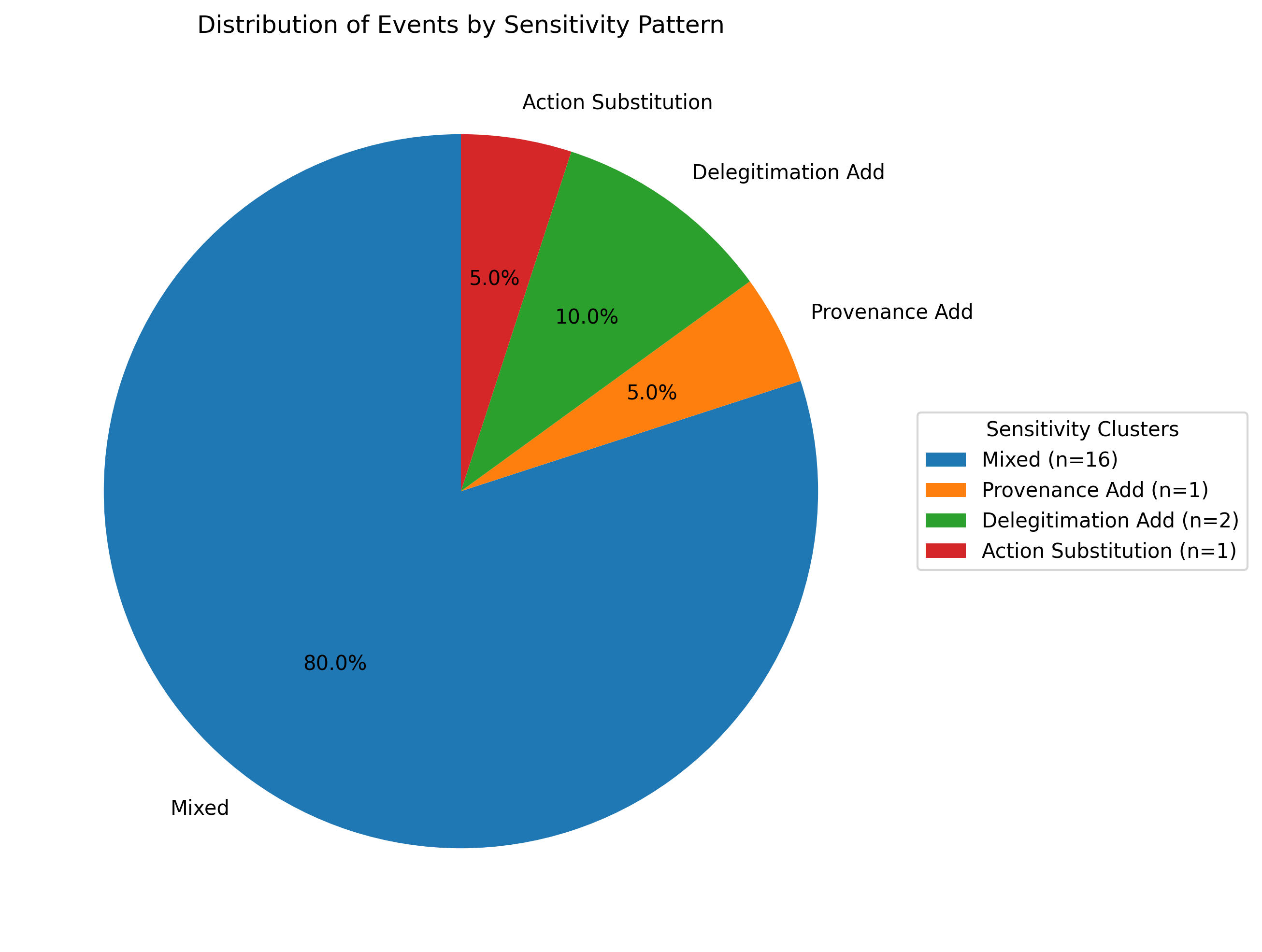}
        \caption{Gemma}
        \label{fig:sensitivity_clusters_gemma}
    \end{subfigure}
    \hfill
    \begin{subfigure}{0.49\textwidth}
        \centering
        \includegraphics[width=\linewidth]{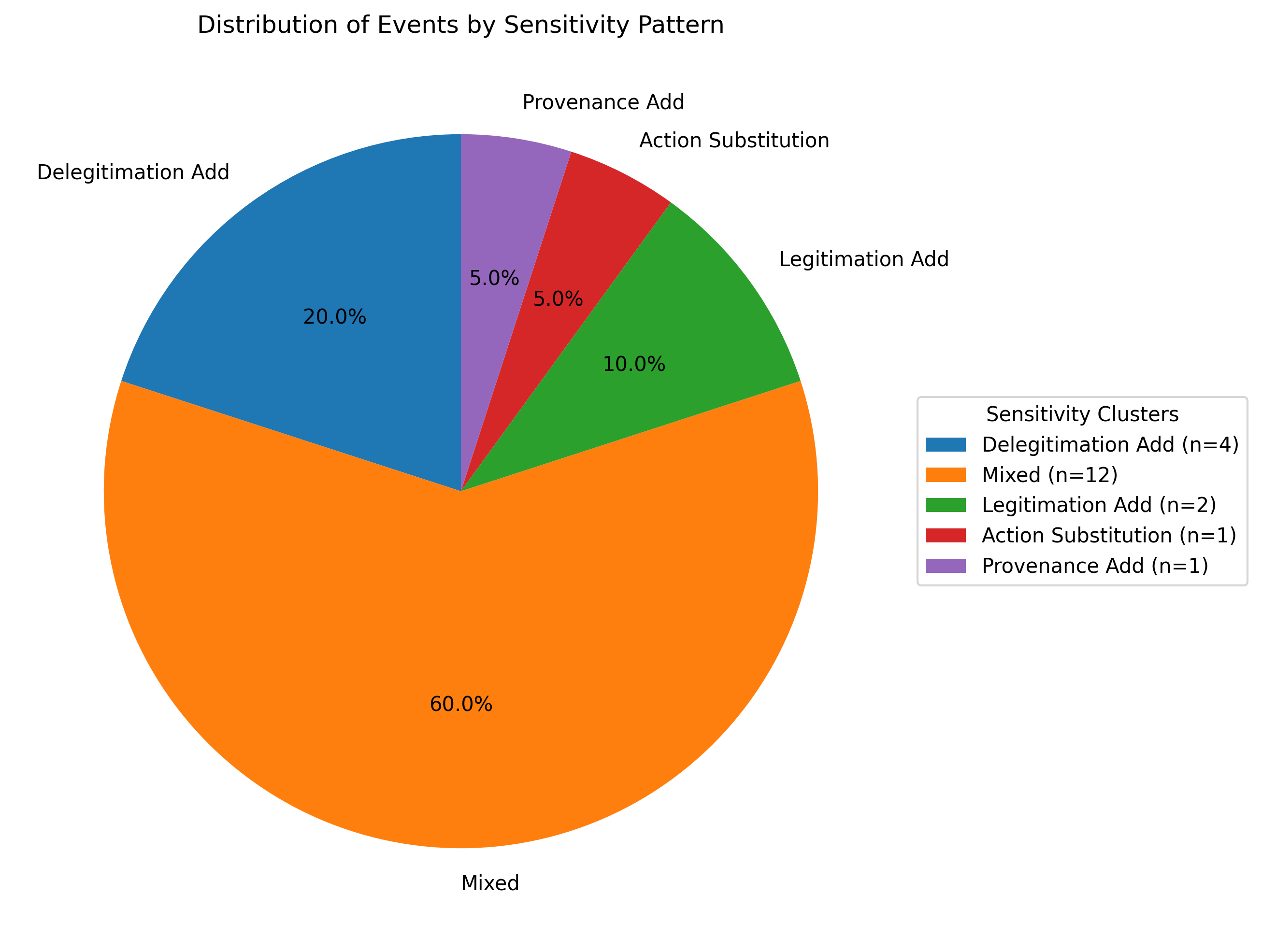}
        \caption{Mistral}
        \label{fig:sensitivity_clusters_mistral}
    \end{subfigure}

    \caption{Sensitivity cluster visualizations for AfroConfliBERT, Olmo, Gemma and Mistral during zero-shot runs on Cameroon. Each plot groups events based on similarity in prediction sensitivity to semantic perturbations. As seen above, different models have different sensitivity clusters.}
    \Description[Sensitivity cluster visualizations for AfroConfliBERT, Olmo, Gemma and Mistral during zero-shot runs on Cameroon]{Sensitivity cluster visualizations for AfroConfliBERT, Olmo, Gemma and Mistral during zero-shot runs on Cameroon. Each plot groups events based on similarity in prediction sensitivity to semantic perturbations. As seen above, different models have different sensitivity clusters.}
    \label{fig:sensitivity_clusters}
\end{figure*}

\end{document}